\documentclass[letterpaper]{article}
\PassOptionsToPackage{table}{xcolor}
\usepackage[preprint]{aaai2027}
\usepackage[hyphens]{url}
\usepackage{graphicx}
\usepackage{natbib}
\usepackage{caption}
\usepackage{booktabs}
\usepackage{amsmath}
\usepackage{amssymb}
\usepackage{multirow}
\title{DepthTile: Depth-Guided Existing-Image Self-Tiling with Input-Side Tileability Assessment}
\author{
Rui Wang\textsuperscript{\rm 1},
Xi Chen\textsuperscript{\rm 1},
Hui Dai\textsuperscript{\rm 1},
Piaopiao Yu\textsuperscript{\rm 2},
Yixue Hao\textsuperscript{\rm 1},
Long Hu\textsuperscript{\rm 1}\corresponding,
Min Chen\textsuperscript{\rm 3}
}
\affiliations{
\textsuperscript{\rm 1}School of Computer Science and Technology,\\
Huazhong University of Science and Technology, Wuhan, China\\
\textsuperscript{\rm 2}Nanjing University of Aeronautics and Astronautics\\
\textsuperscript{\rm 3}School of Computer Science and Engineering, South China University of Technology\\
hulong@hust.edu.cn
}
\begin{document}

\maketitle

\begin{abstract}
Existing-image self-tiling aims to transform a given image into a seamlessly repeatable tile while preserving its original semantics and local details. This task requires weak editing to retain the source but leave periodic seams, whereas strong editing improves tileability at the large risk of content drift.
Existing diffusion-based tiling pipelines alleviate boundary discontinuities, yet lack awareness of which regions are geometrically important to preserve.

In this work, we propose \textbf{DepthTile}, a depth-guided self-tiling method whose generation module requires no retraining of the diffusion backbone. It converts monocular depth into a spatial preservation prior and integrates it into the timestep-dependent masking mechanism of Tiled Diffusion.
DepthTile converts depth into a spatial editability map and fuses it with the differential diffusion mask.

This strengthens source preservation in near-field regions indicated by the monocular depth prior while retaining editing flexibility near compatible background boundaries.

To access when self-tiling is feasible, we introduce the \textbf{Self-Tiling Ability Score} (STAS), a lightweight input-side predictor based on RGB, gradient, depth, and depth-gradient cues. On a 335-image benchmark,
relative to the default Tiled Diffusion pipeline, DepthTile reduces MAE, LPIPS, and Tiling Score by 12.8\%, 26.7\%, and 8.4\%, respectively.

STAS reliably enables difficulty-aware analysis and further supports compatibility-aware pairwise image tiling.
\end{abstract}

\section{Introduction}

Image tiling produces visual content that can be repeated over an unbounded plane with imperceptible seams and is widely used in texture design, game-asset creation, and panoramic content synthesis \citep{efros1999texture,efros2001quilting,kwatra2003graphcut,gatys2015texture}. Conventional tileable texture synthesis requires a plausible periodic pattern, while existing-image self-tiling needs to produce a repeatable tile while preserving its original semantics and local details.

Image tiling and texture synthesis have been studied through classical pixel sampling, wavelet statistics, and joint texture optimization methods \cite{efros1999texture,kwatra2003graphcut,kwatra2005texture}, followed by neural texture models based on feature statistics, feed-forward generation, and adversarial expansion \cite{gatys2015texture,ulyanov2016texture,zhou2018nonstationary}.
More recently, diffusion-based approaches improve tileable or large-image generation by coordinating overlapping generation paths, synchronizing denoising paths, enforcing cross-view consistency, or exchanging latent context across periodic boundaries \cite{bartal2023multidiffusion,zhang2023diffcollage,lee2023syncdiffusion,tang2023mvdiffusion,frolov2025spotdiffusion,madar2025tiled}.

These methods improve periodic consistency or large-image coherence, but most of them synthesize new content rather than preserve the fine-grained content of a specific image.

\begin{figure}[t]
\centering
\includegraphics[width=\linewidth]{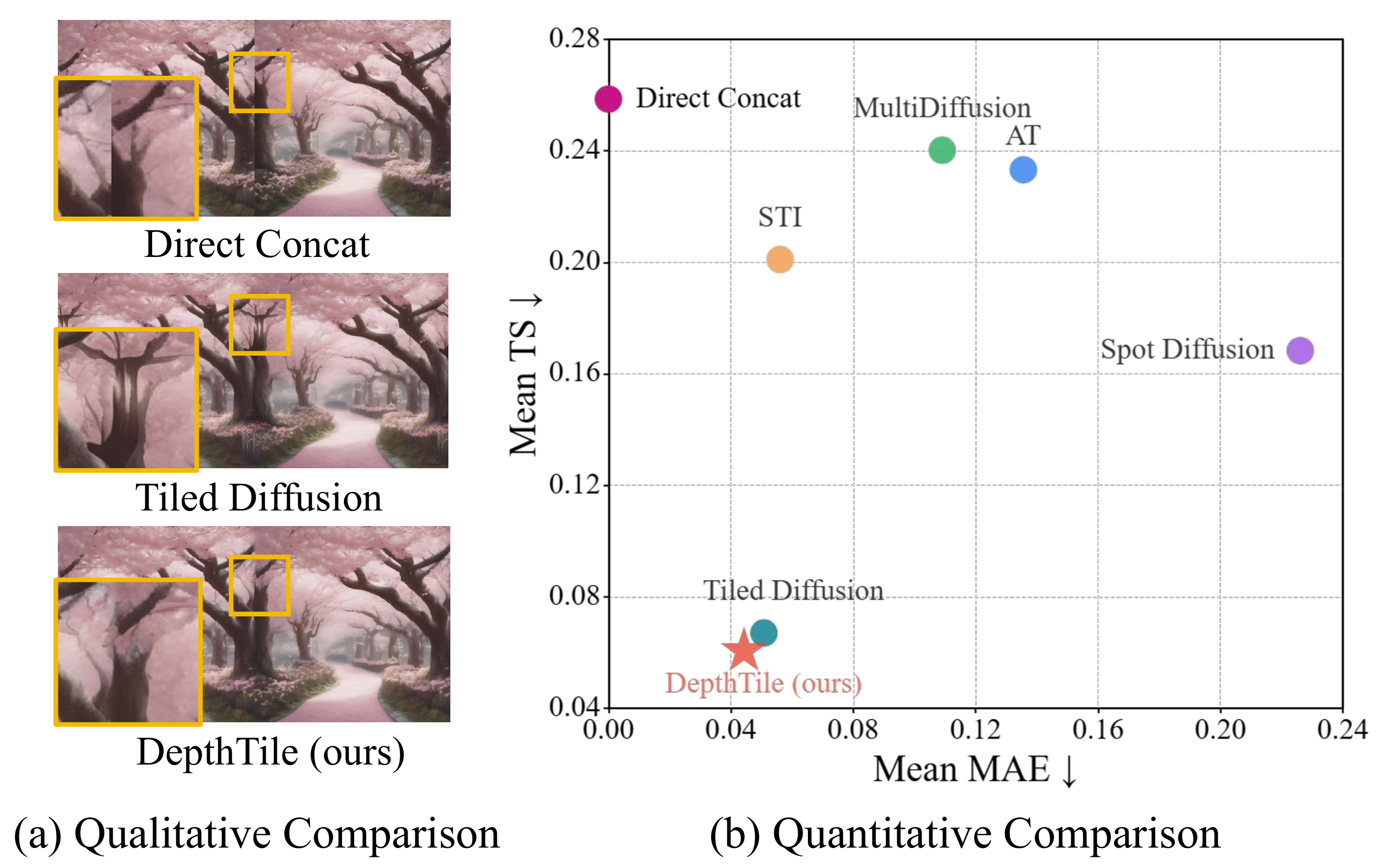}
\caption{Qualitative and quantitative illustration of the preservation--tileability trade-off. (a) Direct concatenation, Tiled Diffusion \citep{madar2025tiled}, and DepthTile on a representative input. (b) Mean MAE and Tiling Score of all compared methods. DepthTile preserves local structures while achieving the lowest mean MAE and Tiling Score.}
\label{fig1:motivation}
\end{figure}

As illustrated in Figure~\ref{fig1:motivation}, existing-image self-tiling is governed by an inherent preservation-tileability trade-off. Direct repetition preserves the source image exactly, but results in visible seams due to discontinuities in texture, contours, or semantic structures.
Diffusion-based tiling can reduce these discontinuities, but stronger regeneration may alter foreground objects, distort near-field structures, or introduce content absent from the source. The central challenge is consequently not only to enforce periodic boundaries, but also to determine where the source should remain protected and where editing freedom should be allocated.
Tiled Diffusion establishes cross-boundary coherence through latent context exchange \citep{madar2025tiled}, whereas differential diffusion assigns spatially varying image-to-image editing strengths through a continuous change map \citep{levin2024differential}. Their combination provides a natural basis for preservation-sensitive self-tiling.
However, a coordinate-defined protection mask is insensitive to the semantic and geometric importance of image regions.
Consequently, seam correction may over-edit foreground or near-field structures while treating them similarly to editable background areas.

Monocular depth provides a natural way to reason about editability because it reveals the spatial organization of a scene.

Near-field regions often contain foreground objects or geometry-sensitive structures that require stronger preservation, while distant background boundaries offer more flexibility for seam adaptation.
Depth has been widely used in image generation as a structural condition, e.g., dedicated control branches \citep{zhang2023controlnet}, lightweight adapters \citep{mou2024t2iadapter}, composable conditioning signals \citep{huang2023composer}, or joint RGB-D diffusion models \citep{wu2024panodiffusion}.
In contrast, existing-image self-tiling benefits from depth as an input-side editability cue to help identify which regions should be protected and which boundary areas can be adapted for seamless repetition.

Motivated by this observation, we propose \textbf{DepthTile}, a training-free framework that incorporates depth-guided protection into diffusion-based self-tiling. DepthTile converts monocular depth into a spatial protection prior and integrates it into the timestep-dependent preservation mechanism used during tiled denoising. The resulting geometry-aware masks protect near and structurally salient regions while retaining the boundary flexibility required for periodic adaptation, without modifying the tiling constraint or retraining the diffusion backbone.
We further observe that self-tiling success is strongly input-dependent: images with compatible boundary appearance, gradients, and depth profiles are naturally easier to tile than images with truncated foreground objects or pronounced boundary conflicts. To capture this intrinsic property, we introduce the \textbf{Self-Tiling Ability Score} (STAS), a lightweight input-side predictor based on RGB, gradient, depth, and depth-gradient boundary cues. STAS supports difficulty analysis, ranking, and candidate selection. On a 335-image benchmark, DepthTile consistently improves the preservation-tileability trade-off over advanced methods. Meanwhile, STAS achieves a Spearman correlation of 0.7184 with the self-tiling ability target and extends naturally to pairwise boundary compatibility.

In summary, we make the following contributions:
\begin{itemize}
    \item We propose DepthTile, a training-free depth-guided framework that preserves important regions while retaining boundary flexibility for existing-image self-tiling.
    \item We design STAS, an input-side tileability predictor that estimates self-tiling difficulty from RGB and depth boundary cues before generation.
    \item Extensive experiments show that DepthTile significantly improves preservation-tileability balance and can extend the boundary representation to pairwise image tiling.

\end{itemize}

\section{Related Work}

\subsection{Image Tiling and Seamless Synthesis}
Classical tiling methods reduce boundary artifacts by stitching overlapping patches or optimizing graph-cut seams \citep{efros2001quilting,kwatra2003graphcut}. Although effective for stationary textures, these local objectives lack semantic and geometric awareness and may cut through salient structures across incompatible boundaries. Diffusion-based methods improve large-content coherence by coordinating overlapping denoising processes \citep{bartal2023multidiffusion,zhang2023diffcollage,lee2023syncdiffusion,tang2023mvdiffusion,frolov2025spotdiffusion}. However, they mainly synthesize new content, allowing substantial deviation from a source image. Tiled Diffusion exchanges latent context between connected boundaries \citep{madar2025tiled}, but its existing-image setting uses a coordinate-defined protection profile that cannot distinguish foreground structures from editable background regions. Thus, current pipelines still lack a geometry-aware mechanism for deciding which source structures should remain unchanged.

\subsection{Diffusion-Based Image Editing}
Diffusion editing methods preserve source information through input noising, spatial masks, semantic instructions, attention control, feature injection, or inversion \citep{meng2022sdedit,lugmayr2022repaint,avrahami2022blended,couairon2023diffedit,brooks2023instructpix2pix,hertz2023prompttoprompt,tumanyan2023pnp,mokady2023nulltext}. These approaches are designed mainly for prompt-directed edits and provide global or region-level control rather than the periodic boundary adaptation required by self-tiling. Differential Diffusion converts a continuous spatial map into nested timestep masks \citep{levin2024differential}, but assumes that a suitable change map is already available. It does not determine which structures should remain protected when seam correction competes with source preservation, motivating a task-specific prior that can be derived automatically from the input.

In the related task of image morphing, AlignMorph uses explicit semantic transport for tuning-free diffusion-based transitions between two input images \citep{liu2026alignmorph}.

\subsection{Depth-Guided Generation and Editing}
Depth-conditioned systems use control networks, adapters, compositional conditions, or joint RGB-D diffusion to specify the geometry of synthesized content \citep{zhang2023controlnet,mou2024t2iadapter,huang2023composer,wu2024panodiffusion}. These methods train dedicated conditioning or generation components and use depth to control what geometry is produced, rather than how strongly an existing image should be preserved during localized boundary repair. Monocular estimators such as MiDaS, ZoeDepth, and Depth Anything V2 recover geometry from a single image, making depth readily available as an input-side prior for preservation-aware editing\citep{ranftl2021vision,ranftl2022midas,bhat2023zoedepth,yang2024depthanythingv2}.

\section{Preliminaries}

\begin{figure*}[t]
\centering
\includegraphics[width=\textwidth]{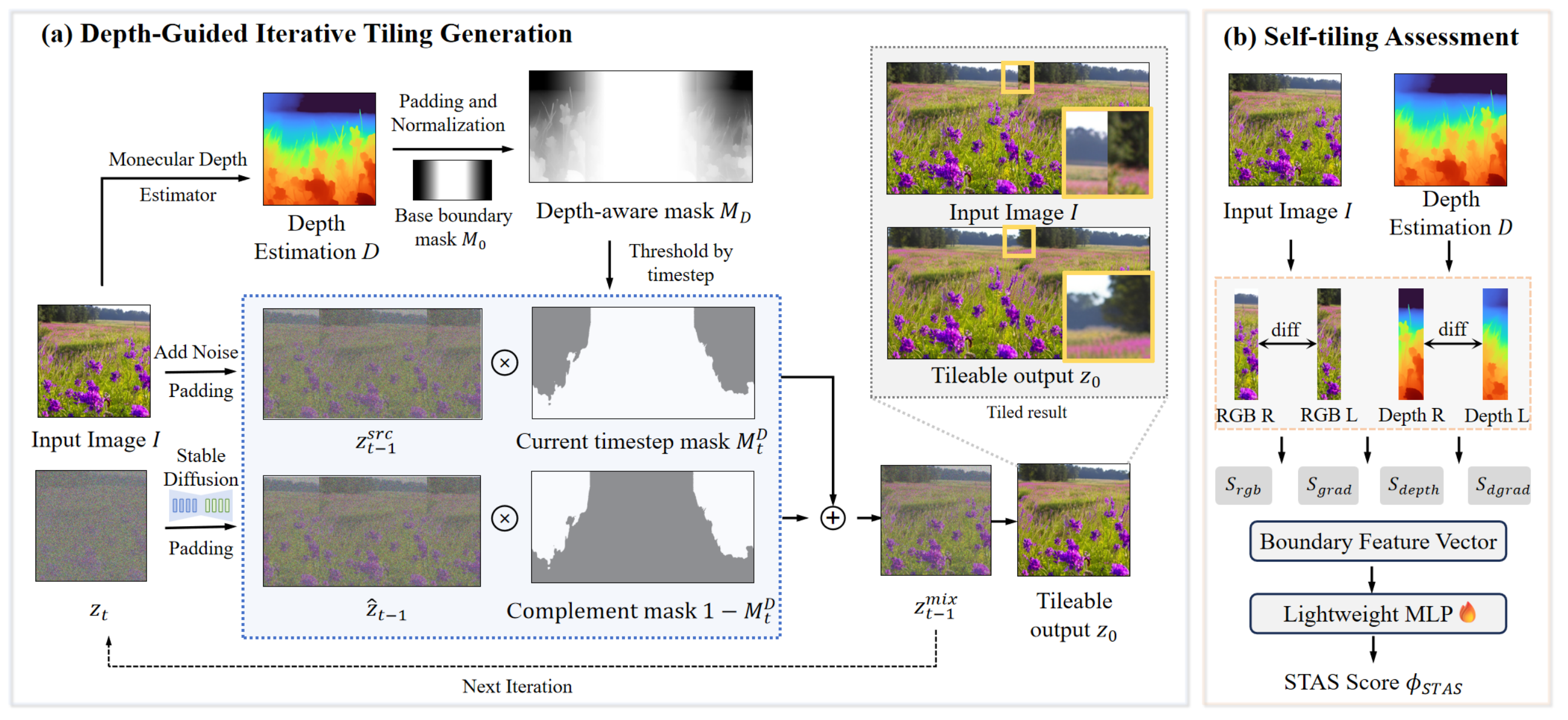}
\caption{Overview of DepthTile. (a) Depth-guided iterative tiling generation converts monocular depth into timestep-dependent masks that mix source and generated latents; the inset compares the input with the tileable output. (b) Self-tiling assessment encodes RGB and depth boundary cues with a lightweight MLP to predict $\phi_{\mathrm{STAS}}$.}
\label{fig:overview}
\end{figure*}

\subsection{Existing-Image Self-Tiling}
Given an input image $I\in\mathbb{R}^{H\times W\times 3}$, existing-image self-tiling aims to produce an edited image $\tilde{I}$ that can be periodically repeated without visible seams while preserving the semantic content and local details of $I$. In this work, we focus on horizontal self-tiling and denote the left and right boundary strips of width $w$ by $B_w^{L}(I)$ and $B_w^{R}(I)$, respectively. The current implementation and evaluation are limited to the horizontal direction; requirements for a controlled vertical extension are summarized in Appendix~\ref{app:vertical_status}.

For horizontal self-tiling, seam inconsistency is measured by comparing the right and left boundaries of the image:
\begin{equation}
\mathcal{L}_{\mathrm{tile}}(\tilde{I})
=
d_{\mathrm{seam}}
\left(
B_w^{R}(\tilde{I}),
B_w^{L}(\tilde{I})
\right),
\end{equation}
where $d_{\mathrm{seam}}(\cdot,\cdot)$ measures cross-boundary discrepancy in pixel, gradient, structural, or perceptual feature spaces.
Unlike unconstrained tileable texture synthesis, existing-image self-tiling is required to preserve the input image. The preservation loss is thus defined as
\begin{equation}
\mathcal{L}_{\mathrm{pres}}(\tilde{I},I)
=
d_{\mathrm{pres}}(\tilde{I},I),
\end{equation}
where $d_{\mathrm{pres}}(\cdot,\cdot)$ is a measurement such as low-level reconstruction distances, structural similarity, or perceptual metrics. Conceptually, the existing-image self-tiling can be formulated as the following bi-objective optimization problem:
\begin{equation}
\tilde{I}^{*}
=
\arg\min_{\tilde{I}}
\;
\mathcal{L}_{\mathrm{tile}}(\tilde{I})
+
\lambda_{\mathrm{pres}}
\mathcal{L}_{\mathrm{pres}}(\tilde{I},I),
\end{equation}
where $\lambda_{\mathrm{pres}}$ controls the balance between periodic continuity and source preservation. This formulation aims to obtain a tiled image $\tilde{I}^{*}$ and resolves the preservation-tileability trade-off: reducing boundary seams requires stronger editing on the image, whereas preserving the source image limits the freedom available for periodic adaptation.

\subsection{Differential Diffusion}
Differential Diffusion controls editing strength through a timestep-dependent preservation mask. Let $z_{t-1}^{\mathrm{src}}$ denote the encoded input noised to timestep $t-1$, and let $\hat{z}_{t-1}$ denote the predicted latent from the current generated trajectory. The two trajectories are combined as
\begin{equation}
z_{t-1}
=
M_t\odot z_{t-1}^{\mathrm{src}}
+
(1-M_t)\odot\hat{z}_{t-1},
\end{equation}
where $M_t\in[0,1]^{h\times w}$ is the preservation mask and $\odot$ denotes element-wise multiplication over spatial locations. Larger mask values inject more source information, whereas smaller values permit stronger editing.

The mask sequence is obtained by thresholding a continuous protection map $M\in[0,1]^{h\times w}$, defined as:
\begin{equation}
M_t(p)
=
\mathbb{I}\!\left[M(p)\geq\tau_t\right],
\end{equation}
where $p$ indexes a spatial location, $\tau_t$ is a monotonic timestep-dependent threshold, and $\mathbb{I}[\cdot]$ is the indicator function. This nested masking mechanism provides spatially varying editing strength without retraining. DepthTile constructs $M$ from depth while keeping the denoising rule unchanged; the exact threshold schedule and mix--wrap--denoise order are given in Appendix~\ref{app:latent_mask}.

\section{Method}

\subsection{Overview of DepthTile}
As shown in Figure~\ref{fig:overview}, DepthTile comprises {depth-guided iterative tiling generation} and {self-tiling assessment}, both using the monocular depth map $D$ estimated from the input image $I$.
{Depth-guided iterative tiling generation} aligns and periodically pads the depth prior before fusing it with the base boundary mask. The resulting protection map is thresholded into timestep-dependent masks that mix the noised source latent with the generated latent throughout tiled denoising. The final latent is decoded into a tileable output, localizing the required changes while preserving most source content.
{Self-tiling assessment} predicts input difficulty before generation. It compares the left and right RGB and depth boundaries using appearance, gradient, depth, and depth-gradient cues. A lightweight MLP maps the feature vector to $\phi_{\mathrm{STAS}}$, which supports difficulty analysis and input ranking before inference. This score is independent of depth-fusion strength and does not alter the generation trajectory.

\subsection{Depth-Guided Iterative Tiling Generation}
\label{sec:depth_guided_inference}

\paragraph{Depth-guided mask generation.}
Given the input image $I$, a pretrained monocular depth estimator produces a relative depth map $D$. We first align its direction and construct the depth prior, defined as
\begin{equation}
P_D
=
\left[
\operatorname{Norm}
\left(
\operatorname{Inv}(D)
\right)
\right]^{\gamma},
\label{eq:depth_prior}
\end{equation}
where $\operatorname{Inv}(\cdot)$ aligns the depth direction so that larger responses indicate closer regions, $\operatorname{Norm}(\cdot)$ normalizes the aligned map to $[0,1]$, and $\gamma>0$ controls the spatial concentration of the prior. Estimator-specific direction handling, non-finite replacement, normalization, and periodic padding are specified in Appendix~\ref{app:depth_preprocessing}; larger values in $P_D$ receive stronger source preservation.

For horizontal self-tiling, the right boundary strip of $P_D$ is copied to the left padding and the left boundary strip is copied to the right padding. The resulting periodically aligned prior is denoted by $P_D^{\mathrm{pad}}$. Let $M_0\in[0,1]^{h\times w}$ be the base boundary mask at the latent resolution $h\times w$. We fuse the depth prior with $M_0$ using the screen operation
\begin{equation}
M_D
=
1-
(1-M_0)
\left(
1-\alpha_D P_D^{\mathrm{pad}}
\right),
\label{eq:depth_mask_fusion}
\end{equation}
where $M_D\in[0,1]^{h\times w}$ is the depth-aware protection mask and $\alpha_D\in[0,1]$ controls the contribution of monocular depth. Setting $\alpha_D=0$ recovers the base mask, whereas $\alpha_D=1$ applies the full depth prior. The screen formulation guarantees $M_D\geq M_0$, so depth guidance never lowers the base preservation strength. Image-to-latent resizing and timestep thresholding are detailed in Appendix~\ref{app:latent_mask}.

Following the nested masking mechanism described in Section~3, the continuous mask $M_D$ is converted into a sequence of timestep-dependent masks:
\begin{equation}
M_t^D(p)
=
\mathbb{I}
\left[
M_D(p)\geq \tau_t
\right],
\qquad
t\in\{1,\ldots,T\},
\label{eq:depth_timestep_mask}
\end{equation}
where $p$ indexes a spatial location, $\tau_t$ is the monotonic threshold associated with timestep $t$, $T$ is the total number of denoising steps, and $\mathbb{I}[\cdot]$ is the indicator function. The sequence $\{M_t^D\}_{t=1}^{T}$ determines how long each spatial region remains anchored to the noised source trajectory.

\paragraph{Iterative tiling generation.}
During iterative generation, the source branch and the evolving generated branch are periodically padded so that opposite horizontal boundaries share latent context during denoising. At timestep $t$, one denoising step predicts the generated latent $\hat{z}_{t-1}$ from the current latent $z_t$. The prediction is then combined with the corresponding noised source latent $z_{t-1}^{\mathrm{src}}$ using the current depth-guided mask:
\begin{equation}
z_{t-1}^{\mathrm{mix}}
=
M_t^D\odot z_{t-1}^{\mathrm{src}}
+
\left(1-M_t^D\right)\odot \hat{z}_{t-1},
\label{eq:depth_guided_mixing}
\end{equation}
where $z_{t-1}^{\mathrm{mix}}$ is the latent passed to the next reverse-diffusion step, $z_{t-1}^{\mathrm{src}}$ is obtained by noising the encoded input image to timestep $t-1$, $\hat{z}_{t-1}$ is the latent predicted by the diffusion denoising step, and $\odot$ denotes element-wise multiplication with broadcasting over latent channels. The mask $M_t^D$ controls source preservation, whereas periodic padding and boundary-context exchange enforce cross-boundary continuity. After the final timestep, the latent is decoded and cropped to the original image extent to obtain the tileable output.

\subsection{Self-Tiling Ability Score}

\begin{table*}[t]
\centering
\small
\begin{tabular}{lccccccc}
\toprule
Method
& MAE$\downarrow$
& RMSE$\downarrow$
& SSIM$\uparrow$
& LPIPS$\downarrow$
& DISTS$\downarrow$
& TS$\downarrow$
& GSE$\downarrow$ \\
\midrule
Asymmetric Tiling SD \citep{tjm35asymmetrictiling}
& 0.1356 & 0.1926 & 0.3278 & 0.4562 & 0.2232 & 0.2332 & 0.3341 \\
Seamless Tile Inpainting \citep{brick2face2023seamlesstile}
& 0.0561 & 0.1153 & \textbf{0.7519} & 0.1173 & \underline{0.0771} & 0.2010 & 0.3261 \\
MultiDiffusion \citep{bartal2023multidiffusion}
& 0.1091 & 0.1606 & 0.4036 & 0.3683 & 0.1831 & 0.2402 & 0.3173 \\
Spot Diffusion \citep{frolov2025spotdiffusion}
& 0.2261 & 0.3078 & 0.2519 & 0.5842 & 0.1686 & 0.1683 & \textbf{0.2797} \\
Tiled Diffusion \citep{madar2025tiled}
& \underline{0.0508} & \underline{0.0856} & 0.6622 & \underline{0.1134} & 0.0804 & \underline{0.0670} & 0.3464 \\
DepthTile (ours)
& \textbf{0.0443} & \textbf{0.0740} & \underline{0.6976} & \textbf{0.0831} & \textbf{0.0648} & \textbf{0.0614} & \underline{0.3074} \\
\bottomrule
\end{tabular}
\caption{Quantitative comparison of existing-image self-tiling methods on the 335-image benchmark. We report content-preservation and seam-quality metrics. Best and second-best results are shown in \textbf{bold} and \underline{underlined}, respectively.}
\label{tab:main_comparison}
\end{table*}

\begin{table*}[t]
\centering
\small
\begin{tabular}{lccccccc}
\toprule
Mask Prior
& MAE$\downarrow$
& RMSE$\downarrow$
& SSIM$\uparrow$
& LPIPS$\downarrow$
& DISTS$\downarrow$
& TS$\downarrow$
& GSE$\downarrow$ \\
\midrule
Tiled Diffusion
& \underline{0.0508} & \underline{0.0856} & \underline{0.6622} & \underline{0.1134} & \underline{0.0804} & 0.0670 & \underline{0.3464} \\
Uniform
& 0.1631 & 0.2255 & 0.2665 & 0.6396 & 0.3778 & 0.0765 & 0.5627 \\
Edge-only
& 0.1302 & 0.1986 & 0.4102 & 0.4003 & 0.2575 & 0.0757 & 0.5064 \\
Saliency-only
& 0.3724 & 0.4560 & 0.1291 & 0.7938 & 0.4480 & 0.0816 & 0.8995 \\
Depth-only
& 0.0927 & 0.1468 & 0.4923 & 0.3206 & 0.2029 & \underline{0.0641} & 0.3641 \\
DepthTile (ours)
& \textbf{0.0443} & \textbf{0.0740} & \textbf{0.6976} & \textbf{0.0831} & \textbf{0.0648} & \textbf{0.0614} & \textbf{0.3074} \\
\bottomrule
\end{tabular}
\caption{Comparison of spatial mask priors under an otherwise identical Tiled Diffusion pipeline, isolating uniform, edge, saliency, and depth cues. Best and second-best results are shown in \textbf{bold} and \underline{underlined}, respectively.}
\label{tab:mask_comparison}
\end{table*}

As shown in Figure~\ref{fig:overview}, STAS quantifies the intrinsic compatibility of the input boundaries before image generation. For horizontal self-tiling, we define
\begin{equation}
\phi_{\mathrm{STAS}}(I)\in[0,1],
\label{eq:stas_definition}
\end{equation}
where $\phi_{\mathrm{STAS}}(I)$ denotes the predicted self-tiling suitability of input image $I$, and a larger value indicates stronger intrinsic tileability.

Using the left and right boundary strips defined in Section~3, we compute four boundary mismatch features: RGB $S_{\mathrm{rgb}}$, image gradient $S_{\mathrm{grad}}$, depth $S_{\mathrm{depth}}$, and depth gradient $S_{\mathrm{dgrad}}$. Their Sobel operators, 32-pixel strips, normalization, and directional forms are specified in Appendix~\ref{app:stas_features}. These features are concatenated into
\begin{equation}
\mathbf{f}(I)
=
\left[
S_{\mathrm{rgb}},
S_{\mathrm{grad}},
S_{\mathrm{depth}},
S_{\mathrm{dgrad}}
\right]^{\mathsf{T}},
\label{eq:stas_feature}
\end{equation}
where $\mathbf{f}(I)\in\mathbb{R}^{4}$ is the boundary feature vector of image $I$. The four components measure mismatch in boundary appearance, local image structure, relative depth, and geometric transitions, respectively.

A lightweight multilayer perceptron (MLP) transforms the boundary representation into STAS:
\begin{equation}
\phi_{\mathrm{STAS}}(I)
=
\sigma
\left(
g_{\psi}
\left(
\mathbf{f}(I)
\right)
\right),
\label{eq:stas}
\end{equation}
where $g_{\psi}$ denotes the MLP parameterized by $\psi$, and $\sigma(\cdot)$ is the logistic sigmoid function. The regression target is a pseudo label constructed from normalized TS, MAE, SSIM, and LPIPS of completed self-tiling results. Its formula, train-only normalization, MLP architecture, and optimization are provided in Appendix~\ref{app:stas_training}. Once trained, the MLP estimates suitability before diffusion inference. We use $\phi_{\mathrm{STAS}}$ for difficulty analysis, input ranking, and compatible-pair selection, sorting all 335 images into contiguous Easy, Medium, and Hard subsets of 112, 112, and 111 images.

\subsection{Extension to Pairwise Image Tiling}

\begin{figure*}[t]
\centering
\includegraphics[width=0.84\textwidth]{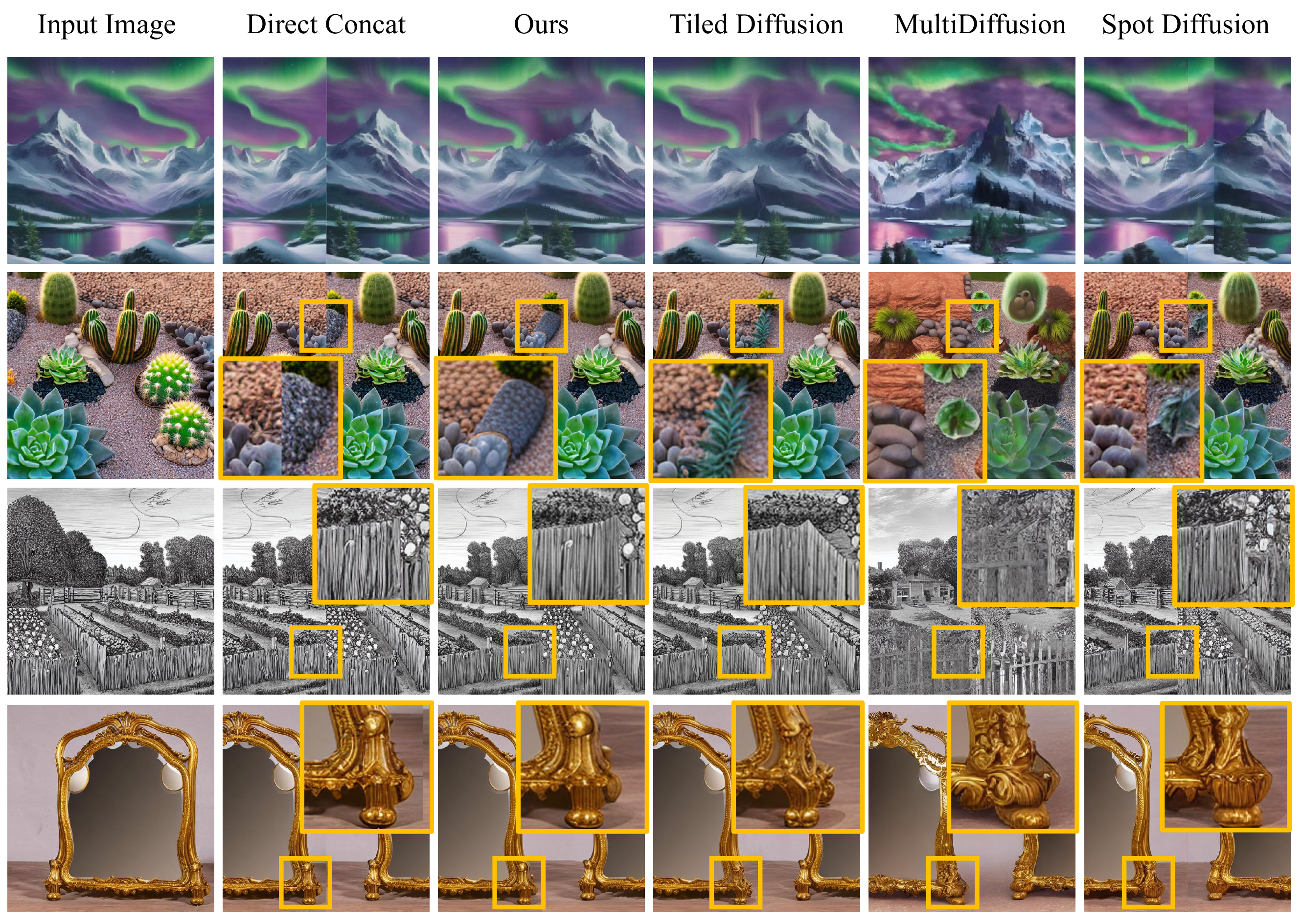}
\caption{Qualitative comparison of existing-image self-tiling methods. Columns show the input and outputs from direct concatenation, DepthTile, Tiled Diffusion, MultiDiffusion, and Spot Diffusion; yellow boxes mark enlarged regions. DepthTile better preserves fine structures while reducing visible tiling artifacts.}
\label{fig:qualitative_comparison}
\end{figure*}

\begin{table*}[t]
\centering
\small
\setlength{\tabcolsep}{4pt}
\begin{tabular}{lcccccccc}
\toprule
Depth Estimator
& Params (M)$\downarrow$
& MAE$\downarrow$
& SSIM$\uparrow$
& LPIPS$\downarrow$
& DISTS$\downarrow$
& TS$\downarrow$
& GSE$\downarrow$
& Time (s)$\downarrow$ \\
\midrule
Depth Anything V2-S
& \textbf{24.8} & 0.0469 & 0.6834 & 0.0941 & 0.0702 & 0.0641 & 0.3091 & \textbf{3.338} \\
MiDaS DPT-Hybrid
& \underline{123.0} & \underline{0.0466} & \underline{0.6851} & \underline{0.0928} & \underline{0.0694} & \underline{0.0634} & \underline{0.3079} & \underline{3.353} \\
ZoeDepth-N
& $\sim$345 & \textbf{0.0443} & \textbf{0.6976} & \textbf{0.0831} & \textbf{0.0648} & \textbf{0.0614} & \textbf{0.3074} & 3.698 \\
\bottomrule
\end{tabular}
\caption{Effect of the monocular depth estimator on DepthTile. Output quality, model size, and inference time are compared under identical tiling settings. Best and second-best results are shown in \textbf{bold} and \underline{underlined}, respectively.}
\label{tab:depth_estimators}
\end{table*}

Pairwise tiling replaces the periodic left--right boundary with a single seam between two inputs, but retains the same need to protect source structures while allowing localized seam correction. We therefore extend DepthTile from cyclic self-tiling to a shared canvas containing two images. Given images $I_A$ and $I_B$, the directional boundary feature vector $\mathbf{f}_{A\rightarrow B}$ is constructed by comparing the right boundary of $I_A$ with the left boundary of $I_B$ using the same RGB, gradient, depth, and depth-gradient cues. Their directional compatibility is
\begin{equation}
\phi_{\mathrm{STAS}}
\left(
I_A\rightarrow I_B
\right)
=
\sigma
\left(
g_{\psi}
\left(
\mathbf{f}_{A\rightarrow B}
\right)
\right),
\label{eq:pairwise_stas}
\end{equation}
where $\phi_{\mathrm{STAS}}(I_A\rightarrow I_B)\in[0,1]$ is the predicted suitability of placing $I_B$ to the right of $I_A$, and $\mathbf{f}_{A\rightarrow B}\in\mathbb{R}^{4}$ is the corresponding directional boundary feature vector. Directional ordering and non-periodic outer-boundary handling are detailed in Appendix~\ref{app:pairwise_details}.

The selected pair is concatenated into the shared canvas $I_{AB}=[I_A\mid I_B]$ and jointly denoised. Unlike cyclic self-tiling, only the center seam is constrained. We define the base preservation profile as
\begin{equation}
M_{\mathrm{base}}(x)
=
m_{\mathrm{seam}}
+
\left(
m_{\mathrm{outer}}-m_{\mathrm{seam}}
\right)
G(x)^{\eta},
\label{eq:pairwise_base_mask}
\end{equation}
where $x$ is the horizontal coordinate on the shared canvas, $G(x)\in[0,1]$ increases with the distance from the center seam, $m_{\mathrm{seam}}$ and $m_{\mathrm{outer}}$ specify the preservation strengths at the seam and outer image regions, respectively, and $\eta>0$ controls the spatial transition between them. This profile leaves more editing freedom near the connection while protecting the interior of both source images. The two depth priors are concatenated and fused with $M_{\mathrm{base}}$ using the screen operation in Eq.~\eqref{eq:depth_mask_fusion}. During joint denoising, latent context from both sides is exchanged across the center seam within the same reverse-diffusion trajectory. This formulation concentrates adaptation near the connection while preserving the interiors of both inputs.

\section{Experiments}

\begin{figure*}[t]
\centering
\begin{minipage}[t]{0.65\textwidth}
\vspace{0pt}
\centering
\scriptsize
\setlength{\tabcolsep}{3pt}
\resizebox{\linewidth}{!}{%
\begin{tabular}{lcccccccc}
\toprule
& \multicolumn{4}{c}{Validation} & \multicolumn{4}{c}{Test} \\
\cmidrule(lr){2-5}
\cmidrule(lr){6-9}
Variant
& MAE$\downarrow$
& RMSE$\downarrow$
& PCC$\uparrow$
& SRCC$\uparrow$
& MAE$\downarrow$
& RMSE$\downarrow$
& PCC$\uparrow$
& SRCC$\uparrow$ \\
\midrule
Full
& \underline{0.0930}
& \textbf{0.1111}
& \textbf{0.2914}
& \textbf{0.3424}
& \textbf{0.0810}
& \textbf{0.1003}
& \textbf{0.6147}
& \textbf{0.7184} \\
Appearance
& 0.0965
& 0.1157
& 0.0823
& 0.0679
& 0.0958
& 0.1236
& 0.2657
& 0.3671 \\
Gradient
& \textbf{0.0902}
& \underline{0.1114}
& \underline{0.2575}
& \underline{0.3149}
& \underline{0.0895}
& \underline{0.1125}
& \underline{0.4687}
& \underline{0.5322} \\
\bottomrule
\end{tabular}
}
\captionof{table}{Ablation of the boundary features used to predict STAS. Full, appearance-only, and gradient-only representations use the same data splits and MLP. Best and second-best results are shown in \textbf{bold} and \underline{underlined}, respectively.}
\label{tab:stas_input_ablation}
\end{minipage}\hfill
\begin{minipage}[t]{0.3\textwidth}
\vspace{0pt}
\centering
\includegraphics[width=\linewidth]{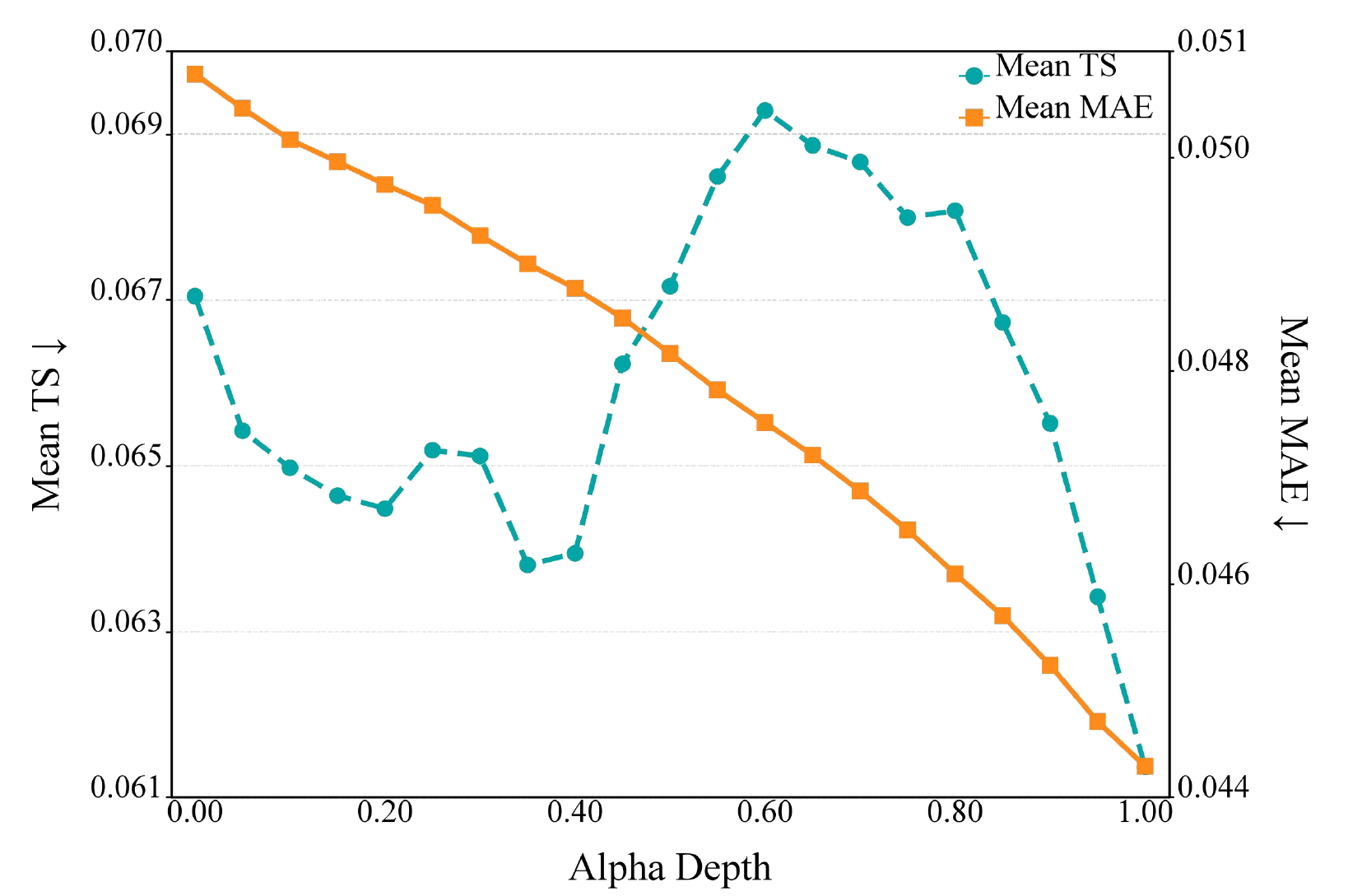}
\captionof{figure}{Effect of depth-fusion strength. Full depth guidance achieves the lowest MAE and Tiling Score.}
\label{fig:alpha_ablation}
\end{minipage}
\end{figure*}

\begin{table*}[t]
\centering
\small
\begin{tabular}{llccccccc}
\toprule
Subset & Method
& MAE$\downarrow$
& RMSE$\downarrow$
& SSIM$\uparrow$
& LPIPS$\downarrow$
& DISTS$\downarrow$
& TS$\downarrow$
& GSE$\downarrow$ \\
\midrule
\multirow{2}{*}{All pairs}
& Tiled Diffusion
& \underline{0.0403} & \underline{0.0725} & \underline{0.7489}
& \underline{0.0862} & \underline{0.0505} & \textbf{0.1349} & \textbf{0.2789} \\
& DepthTile
& \textbf{0.0355} & \textbf{0.0649} & \textbf{0.7722}
& \textbf{0.0638} & \textbf{0.0425} & \underline{0.1621} & \underline{0.2812} \\
\midrule
\multirow{2}{*}{Top-50 STAS}
& Tiled Diffusion
& \underline{0.0305} & \underline{0.0594} & \underline{0.8402}
& \underline{0.0592} & \underline{0.0430} & \textbf{0.0830} & \underline{0.1305} \\
& DepthTile
& \textbf{0.0272} & \textbf{0.0549} & \textbf{0.8532}
& \textbf{0.0438} & \textbf{0.0376} & \underline{0.1019} & \textbf{0.1219} \\
\bottomrule
\end{tabular}
\caption{Quantitative comparison of Tiled Diffusion and DepthTile for pairwise tiling on 100 image pairs and the top-50 pairs selected by Pairwise STAS. Best and second-best results within each subset are shown in \textbf{bold} and \underline{underlined}, respectively.}
\label{tab:pairwise_summary}
\end{table*}

\subsection{Experimental Settings}

\subsubsection{Dataset and protocol}
Following the evaluation protocol of Tiled Diffusion, we sample prompts from LAION-400M \citep{schuhmann2021laion400m} and retain 335 images at $512\times512$ resolution. We evaluate horizontal existing-image self-tiling and visualize each result as a $1\times3$ repetition; all methods use the same source images and evaluation code. Prompt filtering, semantic categories, and the fixed 234/50/51 STAS split are detailed in Appendix~\ref{app:benchmark}. The retained construction note conflicts with the current SDXL and seed-42 record, so the source model and construction seed require reconciliation before release. The pairwise study uses 100 pairs and a top-50 Pairwise-STAS subset; its maintained directional manifest and snapshot limitations are described in Appendix~\ref{app:data_splits}.

\subsubsection{Implementation details}
DepthTile is implemented on top of Tiled Diffusion without retraining the diffusion backbone. Stable Diffusion XL is the generative backbone, and ZoeDepth-N \citep{bhat2023zoedepth} is the default depth estimator. We set $\gamma=0.5$ and $\alpha_D=1.0$, while the ablation evaluates $\alpha_D\in\{0,0.05,\ldots,1.0\}$. Other scheduler, denoising, mask, and latent-tiling settings remain fixed. Frozen alpha/gamma outputs use seed 151, whereas STAS splitting and training use seed 42; complete settings are listed in Appendix~\ref{app:record}. The STAS head has 2,433 parameters and 4,672 linear FLOPs, with 0.109 ms head-only latency; stage-wise runtime and memory are reported in Appendix~\ref{app:efficiency}.

\subsubsection{Metrics}
Content preservation is evaluated using MAE, RMSE, SSIM, LPIPS, and full-image DISTS \citep{wang2004ssim,zhang2018lpips,ding2022dists}, while seam quality uses Tiling Score (TS) and Gradient Seam Error (GSE). Exact image ranges, package settings, formulas, and boundary widths are provided in Appendix~\ref{app:metrics}. Higher SSIM is better; lower values are better otherwise. TS follows Tiled Diffusion and probes the repeated boundary and nearby transitions.

\subsection{Main Results}
\subsubsection{Quantitative comparison results}
We compare DepthTile with adapted self-tiling methods under identical inputs and output resolution. As shown in Table~\ref{tab:main_comparison}, DepthTile reduces MAE from 0.0508 to 0.0443 and TS from 0.0670 to 0.0614. Seamless Tile Inpainting achieves higher SSIM but TS 0.2010, while Spot Diffusion obtains lower GSE by allowing substantially larger content changes.

\subsubsection{Qualitative comparison results}
Figure~\ref{fig:qualitative_comparison} complements the aggregate metrics with representative examples. Direct concatenation retains the source content but leaves incompatible boundaries unresolved, whereas the diffusion baselines may alter fine textures, thin structures, or local geometry. DepthTile more faithfully preserves image details while adapting the image for repetition.

\subsection{Ablation Studies}

\subsubsection{Mask Prior}
Table~\ref{tab:mask_comparison} isolates the effect of spatial priors. DepthTile improves MAE from 0.0508 to 0.0443 and TS from 0.0670 to 0.0614 over Tiled Diffusion. Depth-only achieves a competitive TS of 0.0641 but degrades MAE to 0.0927 because it removes the base boundary profile and over-constrains editable regions. Uniform, edge-only, and saliency-only priors also underperform, confirming the complementarity of depth and the base mask.

\subsubsection{Depth-Fusion Strength}
Figure~\ref{fig:alpha_ablation} evaluates $\alpha_D=0.05k$ for $k=0,\ldots,20$ with $\gamma=0.5$. From $\alpha_D=0$ to 1.0, MAE decreases from 0.0508 to 0.0443 and TS from 0.0670 to 0.0614. Intermediate TS is non-monotonic because partial guidance protects structures inconsistently across the boundary neighborhood. Full guidance gives the best overall trade-off. The complete 21-point sweep is in Appendix~\ref{app:alpha_sweep}, with Depth-Anything-based gamma sensitivity in Appendix~\ref{app:gamma_sensitivity}.

\subsubsection{Depth Estimator Robustness}
Table~\ref{tab:depth_estimators} compares three estimators under identical settings. ZoeDepth-N gives the best MAE (0.0443) and TS (0.0614) but uses about 345M parameters. Depth Anything V2-S uses 24.8M parameters and is faster (3.338\,s versus 3.698\,s), with MAE increasing only to 0.0469. DepthTile is therefore robust to estimator choice, while ZoeDepth-N remains the quality-oriented default.

\subsection{STAS Validation}

\subsubsection{Boundary Feature Ablation}
Table~\ref{tab:stas_input_ablation} compares feature variants using the same split and MLP. Gradient yields the lowest validation MAE (0.0902 versus 0.0930 for Full), but this advantage does not transfer to testing. Full achieves the strongest test correlations, with PCC 0.6147 and SRCC 0.7184, compared with 0.4687 and 0.5322 for Gradient. Appearance performs markedly worse. These results indicate that gradient cues are useful but less stable alone, whereas combining appearance, depth, and gradient information improves generalization.

\subsubsection{Difficulty-Stratified Analysis}
\begin{figure}[h]
\centering
\includegraphics[width=\linewidth]{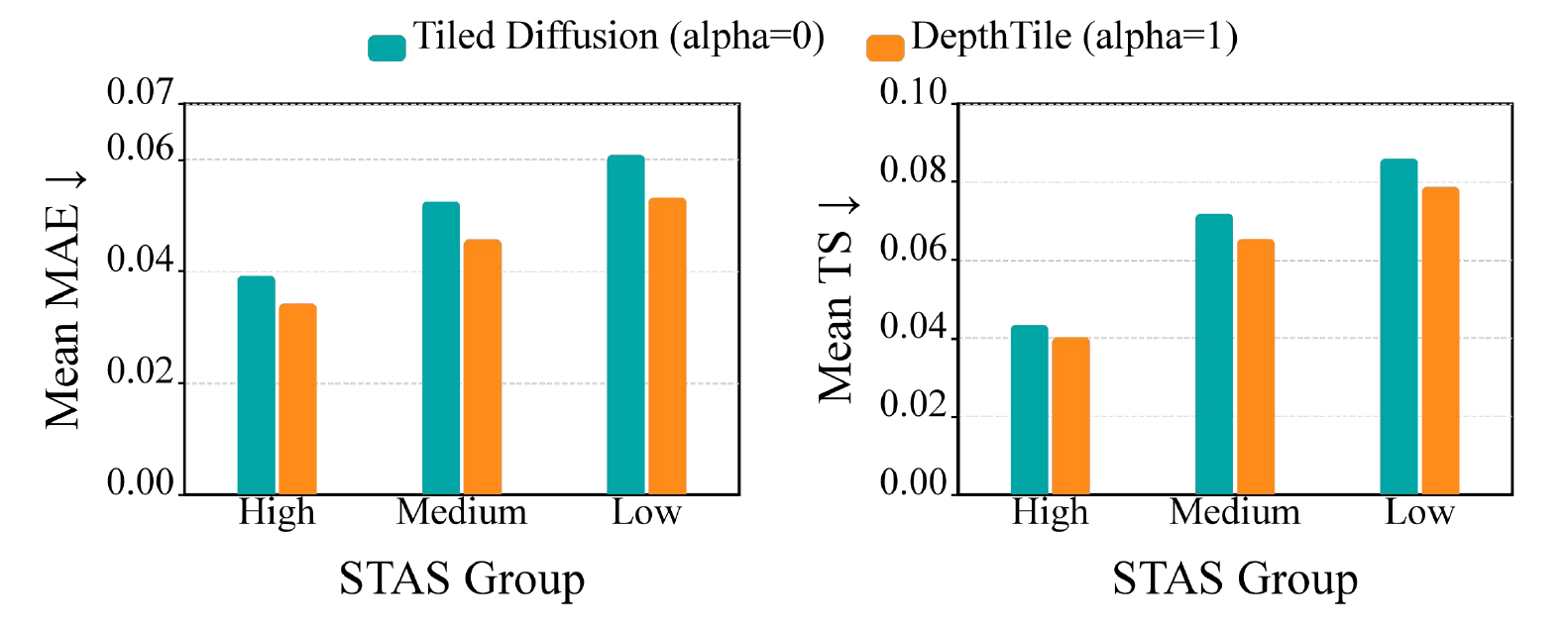}
\caption{STAS-stratified self-tiling performance. DepthTile consistently reduces MAE and TS across all difficulty levels.}
\label{fig:stas_groups}
\end{figure}

Figure~\ref{fig:stas_groups} partitions the STAS ranking into Easy, Medium, and Hard subsets of 112, 112, and 111 images. For Tiled Diffusion, MAE increases from 0.0392 to 0.0608 and TS from 0.0434 to 0.0860 from Easy to Hard, supporting STAS as an input-side difficulty predictor. DepthTile improves all subsets, reducing Hard-set MAE and TS to 0.0531 and 0.0787, although challenging inputs remain intrinsically harder.

\subsection{Pairwise Image Tiling}

Table~\ref{tab:pairwise_summary} evaluates 100 pairs and the top-50 Pairwise STAS subset on pairwise tiling. Across all pairs, DepthTile reduces MAE from 0.0403 to 0.0355 but raises TS from 0.1349 to 0.1621, as stronger preservation limits seam repair for incompatible inputs. Selecting compatible pairs alleviates this trade-off: TS decreases to 0.0830 for Tiled Diffusion and 0.1019 for DepthTile.

\section{Conclusion}
We present DepthTile, a geometry-aware inference framework for existing-image self-tiling. By converting monocular depth into a spatial protection prior, DepthTile reduces unnecessary changes to near-field regions during periodic boundary adaptation. The method improves both fidelity and Tiling Score over the default tiled diffusion pipeline and outperforms existing self-tiling baselines across complementary metrics.
Furthermore, STAS provides an efficient input-side estimate of self-tiling difficulty and consistently separates easy from hard cases before generation. Its boundary representation also supports pairwise compatibility selection.
Our proposed method provides a practical framework for preservation-sensitive image tiling without retraining the diffusion backbone.

\bibliography{references}

\clearpage
\appendix
\twocolumn[{\centering\LARGE\bfseries Appendix\par\vspace{1em}}]

\section{Reproducibility Record}
\label{app:record}

This supplement draws on three sources of evidence: executable
implementations, frozen per-image result files, and the retained benchmark
construction record. All numerical tables are recomputed by aggregating the
frozen per-image results; no diffusion outputs are regenerated for the
supplement. The aggregation script is
\path{scripts/build_appendix_statistics.py}, and the corresponding validation
report is
\path{outputs/appendix_statistics/APPENDIX_STATISTICS_REPORT.md}. Each of the
21 depth-strength settings and eight gamma configurations contains all 335
sample identifiers exactly once, with finite values for the seven reported
metrics.

\begin{table}[t]
\centering
\caption{Reproducible inference configuration used for the horizontal
self-tiling experiments in the ZoeDepth depth-strength sweep.}
\label{tab:appendix_inference_settings}
\begin{tabular}{ll}
\toprule
Component & Setting \\
\midrule
Backbone & SDXL base and refiner, FP16 \\
Depth estimator & ZoeDepth-N \\
Input/output tile & $512\times512$ RGB \\
Scheduler & EulerDiscreteScheduler \\
Denoising steps & 40 \\
Base/refiner split & 0.8 \\
CFG scale & 7.5 \\
Latent context width & 32 (256 image pixels) \\
Image-to-image strength & 1.0 \\
Depth shaping & $\gamma=0.5$ \\
Main depth strength & $\alpha_D=1.0$ \\
Mask fusion & Screen \\
Frozen-output seed & 151 \\
\bottomrule
\end{tabular}
\end{table}

The frozen alpha and gamma outputs use generation seed 151, whereas seed 42
is used only for the STAS data split and MLP training. Reproducing the reported
generation results therefore requires seed 151; the two seeds serve distinct
purposes.

\section{Benchmark and Protocol}
\label{app:benchmark}

\subsection{Prompt Sampling, Classification, and Source Generation}
\label{app:dataset_construction}

The retained construction record specifies the following pipeline. We first
read 5,000 candidate captions from the \path{default/train} split of
\path{laion/relaion400m}. Text is extracted from the first available field
among \path{TEXT}, \path{text}, \path{caption}, \path{description}, and
\path{alt_text}; otherwise, the first nonempty string field is used.
Whitespace is normalized, and the source identifier is retained for every
caption. The recorded positive prompt template is
\begin{quote}
\small\raggedright\ttfamily
original caption, masterpiece, best quality, high resolution, highly
detailed, sharp focus, clear subject, well-composed, realistic lighting,
natural light, balanced exposure, rich texture, clean composition
\end{quote}

DeepSeek V4 Pro assigns each caption to at most one of 12 visual categories
using batches of 100, temperature zero, and JSON-only output. We discard
records with a null or out-of-vocabulary category, malformed output, or an
identifier outside the current batch. Each retained caption is augmented with
a common quality suffix emphasizing detail, composition, natural illumination,
balanced exposure, and texture. With probability 0.60, determined
deterministically from the source identifier and base seed, we additionally
append
\begin{quote}
\small\raggedright\ttfamily
prominent foreground subject, clear foreground, layered depth, strong depth
separation
\end{quote}
A shared negative prompt suppresses blur, low resolution, compression
artifacts, abnormal exposure or color, truncated or repeated subjects,
malformed anatomy, text, watermarks, signatures, borders, and logos. We remove
images with failed saves, identifier mismatches, severe blur, missing subjects,
obvious deformation, text, or watermarks.

\paragraph{Source-generation provenance.}
The retained construction record specifies Stable Diffusion v1.5, 40
denoising steps, CFG 7.5, $512\times512$ PNG output, base seed 12,345, and a
per-image seed obtained by adding the category-file line number to the base
seed. This record differs from the main manuscript, which attributes source
generation to SDXL with seed 42. The available repository contains the final
335 images and their source identifiers, but not the original prompt manifest,
generation metadata, or construction executable. We therefore leave the
source generator and construction seed unresolved rather than selecting
between inconsistent records. These fields must be verified against the
original generation archive before the camera-ready release.

\subsection{Final Categories and Data Splits}
\label{app:data_splits}

The final benchmark contains 335 RGB images at $512\times512$. Table
\ref{tab:appendix_categories} reports category counts recovered from the image
directory and confirmed against every alpha group.

\begin{table}[t]
\centering
\caption{Semantic composition of the 335-image benchmark used in all
horizontal self-tiling experiments.}
\label{tab:appendix_categories}
\begin{tabular}{lrlr}
\toprule
Category & $N$ & Category & $N$ \\
\midrule
Animals & 32 & Architecture & 41 \\
Beach & 26 & Desert & 17 \\
Forest & 24 & Garden & 30 \\
Indoor & 31 & Landscape & 33 \\
Mountain & 29 & Object foreground & 35 \\
Road & 17 & Street & 20 \\
\bottomrule
\end{tabular}
\end{table}

For STAS, a deterministic shuffle with seed 42 produces 234 training, 50
validation, and 51 test images. The assignments are stored by image identifier
and shared by all STAS variants. For difficulty analysis, we sort images by
predicted STAS in descending order and partition the ranking into Easy,
Medium, and Hard groups containing 112, 112, and 111 images, respectively.

The maintained pairwise scorer evaluates all
$\binom{335}{2}=55{,}945$ unordered pairs in both directions. After excluding
the selected direction when any feature lies more than three training
standard deviations from its mean, the pipeline retains the 50 highest-scoring
Pairwise-STAS pairs and 50 score-stratified controls. The pairwise table in the
main manuscript originates from an earlier experimental snapshot whose exact
revision was not retained. Accordingly, it should be treated as a frozen
result rather than as an output exactly regenerated by the maintained
directional pipeline.

\section{DepthTile Implementation}
\label{app:depthtile_implementation}

\subsection{Depth Direction, Normalization, and Mask Construction}
\label{app:depth_preprocessing}

Before depth estimation, each input is center-cropped and resized to
$512\times512$ using Lanczos resampling. Let
$D\in\mathbb{R}^{H\times W}$ denote the raw prediction. Non-finite values
are replaced by the median of the finite pixels; a sample is rejected only
when no finite value is available. We apply neither percentile clipping nor
learned calibration. Per-image normalization is
\begin{equation}
\mathcal{N}(D)=\frac{D-\min D}{\max D-\min D+10^{-6}}.
\label{eq:appendix_depth_norm}
\end{equation}
ZoeDepth predicts metric-like depth, for which larger values indicate more
distant regions. Its near-high protection map is therefore
\begin{equation}
P_0=1-\mathcal{N}(D).
\label{eq:appendix_depth_direction}
\end{equation}
For disparity-like estimators whose outputs are already near-high, we instead
set $P_0=\mathcal{N}(D)$. The shaped continuous prior is
$P_D=\operatorname{clip}(P_0,0,1)^\gamma$.

For horizontal tiling, the image-space prior is periodically extended by half
a tile on each side:
\begin{equation}
P_D^{\mathrm{pad}}=
\left[P_D[:,256{:}512]\ \middle|\ P_D\ \middle|\ P_D[:,0{:}256]\right],
\label{eq:appendix_periodic_depth}
\end{equation}
which yields a $512\times1024$ map. The fixed $2048\times1024$ grayscale
Differential Diffusion mask is resized with Lanczos resampling to
$1024\times512$ and scaled to $M_0\in[0,1]$, where larger values indicate
stronger source preservation. Screen fusion then gives
\begin{equation}
M_D=1-(1-M_0)\left(1-\alpha_D P_D^{\mathrm{pad}}\right).
\label{eq:appendix_screen_fusion}
\end{equation}
Thus $\alpha_D=0$ returns $M_0$ exactly and $M_D\geq M_0$ elementwise.

\subsection{Image-to-Latent Mask Transfer and Denoising Order}
\label{app:latent_mask}

Because SDXL uses a VAE scale factor of eight, the continuous
$512\times1024$ mask is resized to $64\times128$ using Torchvision bilinear
interpolation without antialiasing. Resizing precedes thresholding. For global
denoising iteration $j\in\{0,\ldots,T-1\}$ with $T=40$, the implementation
uses the strict comparison
\begin{equation}
M_j(p)=\mathbb{I}\!\left[M_D^{\mathrm{lat}}(p)>\frac{j}{T}\right].
\label{eq:appendix_threshold}
\end{equation}
The base stage processes global iterations 0--31 and the refiner processes
32--39 through \path{denoising_end=0.8} and
\path{denoising_start=0.8}, preserving the global threshold sequence
$0,1/40,\ldots,39/40$ across both stages.

For $j>0$, each iteration applies source mixing, horizontal wrapping, UNet
prediction, and the scheduler update in the following order:
\begin{align}
z_j^{\mathrm{mix}} &=
M_j\odot z_j^{\mathrm{src}}+(1-M_j)\odot z_j^{\mathrm{gen}}, \\
z_j^{\mathrm{wrap}} &= \operatorname{WrapX}
    (z_j^{\mathrm{mix}},32), \\
z_{j+1}^{\mathrm{gen}} &=
\operatorname{EulerStep}\!\left(
z_j^{\mathrm{wrap}},\epsilon_\theta(z_j^{\mathrm{wrap}},t_j),t_j
\right).
\label{eq:appendix_execution_order}
\end{align}
At the first iteration, the current latent is initialized from the noised
source latent. The 64-column source latent is padded by 32 latent columns on
each side, and \path{wrap_edges_x} copies opposite interior context into the
padding before every UNet call. The side regions are cropped after each SDXL
stage, leaving a single $512\times512$ tile. The $1\times3$ repetition is
created only for visualization and is never provided to the diffusion model.

\subsection{Directional Pairwise Extension}
\label{app:pairwise_details}

For an ordered pair $A\rightarrow B$, the scorer compares the right boundary
of $A$ with the left boundary of $B$; the reverse direction is evaluated
independently. The direction with the higher score is selected, with
$A\rightarrow B$ used to break exact ties. The selected images and depth maps
are concatenated into a $1024\times512$ canvas containing one internal seam.
Periodic self-padding and outer-boundary wrapping are disabled. The base
protection profile is weakest at the center seam and increases toward the
image interiors, while the concatenated depth prior is incorporated through
screen fusion. A joint full-canvas UNet call observes both sides of the seam
within the same denoising trajectory.

\section{Self-Tiling Ability Score}
\label{app:stas}

\subsection{Boundary Feature Definitions}
\label{app:stas_features}

Let $I\in[0,1]^{H\times W\times3}$ be the RGB image and
$P_0\in[0,1]^{H\times W}$ its near-high depth prior before gamma shaping.
The luma image is
\begin{equation}
Y=0.2126I_R+0.7152I_G+0.0722I_B.
\end{equation}
For a scalar map $X$, we compute the $3\times3$ OpenCV Sobel magnitude
\begin{equation}
\mathcal{G}(X)=\sqrt{\operatorname{Sobel}_x(X)^2+
                           \operatorname{Sobel}_y(X)^2}
\end{equation}
and normalizes it as
\begin{equation}
\mathcal{Q}_{95}(X)=
\operatorname{clip}\!\left(\frac{X}{q_{0.95}(X)},0,1\right).
\end{equation}
If the 95th percentile is below $10^{-6}$, the normalized map is zero.
This normalization is computed over the full image before boundary strips are
extracted.

Let $B_L^b$ and $B_R^b$ select the left and right $b=32$ columns. The four
stored features are mean absolute boundary mismatches:
\begin{align}
S_{\mathrm{rgb}} &=
\frac{\|B_L^b(I)-B_R^b(I)\|_1}{3Hb}, \\
S_{\mathrm{grad}} &=
\frac{\|B_L^b(\mathcal{Q}_{95}(\mathcal{G}(Y)))-
          B_R^b(\mathcal{Q}_{95}(\mathcal{G}(Y)))\|_1}{Hb}, \\
S_{\mathrm{depth}} &=
\frac{\|B_L^b(P_0)-B_R^b(P_0)\|_1}{Hb}, \\
S_{\mathrm{dgrad}} &=
\frac{\|B_L^b(\mathcal{Q}_{95}(\mathcal{G}(P_0)))-
          B_R^b(\mathcal{Q}_{95}(\mathcal{G}(P_0)))\|_1}{Hb}.
\label{eq:appendix_stas_features}
\end{align}
Although denoted by $S$, these raw features are mismatch measures, so larger
values indicate less compatible boundaries. The MLP learns their mapping to an
ability score for which larger values indicate easier self-tiling.

Each feature is z-scored using statistics computed only on the training
split. In RGB, gradient, depth, and depth-gradient order, the retained means
are $(0.253969,0.248496,0.168594,0.179061)$ and the population standard
deviations are $(0.093446,0.077962,0.119393,0.081681)$.

For $A\rightarrow B$, Eq. \eqref{eq:appendix_stas_features} is modified only
by replacing each self-comparison with $B_R^b(A)$ versus $B_L^b(B)$.
Reversing the pair therefore generally changes all four features and the
predicted score.

\subsection{Pseudo Target, Network, and Optimization}
\label{app:stas_training}

The STAS target is a pseudo-label derived from completed
$\alpha_D=1$ ZoeDepth self-tiling outputs rather than a human annotation or an
absolute measure of tileability. For metric $m$, we define the train-only
transform
\begin{equation}
C_m(x)=\operatorname{clip}\!\left(
\frac{x-m_{\min}^{\mathrm{train}}}
     {m_{\max}^{\mathrm{train}}-m_{\min}^{\mathrm{train}}+10^{-8}},
0,1\right).
\end{equation}
The fitted ranges are TS $[0.003559,0.227870]$, MAE
$[0.013255,0.099941]$, and LPIPS $[0.029048,0.190996]$. The target used for
all 335 samples is
\begin{equation}
\begin{split}
y_{\mathrm{STAS}}=&
0.45(1-C_{\mathrm{TS}}(\mathrm{TS}))
+0.25(1-C_{\mathrm{MAE}}(\mathrm{MAE})) \\ &
+0.15\,\operatorname{clip}(\mathrm{SSIM},0,1)\\ &
+0.15(1-C_{\mathrm{LPIPS}}(\mathrm{LPIPS})).
\label{eq:appendix_stas_target}
\end{split}
\end{equation}

The head is
\begin{equation}
\begin{split}
    4&\rightarrow\operatorname{Linear}(64)\rightarrow\operatorname{ReLU}
\\ &\rightarrow\operatorname{Dropout}(0.1)\rightarrow\operatorname{Linear}(32)
\\ &\rightarrow\operatorname{ReLU}\rightarrow
\operatorname{Linear}(1)\rightarrow\operatorname{Sigmoid}.
\end{split}
\end{equation}
The head contains 2,433 trainable parameters. We optimize full-batch MSE with
AdamW, a learning rate of $10^{-3}$, weight decay $10^{-4}$, seed 42, and a
maximum of 1,000 epochs. Early stopping monitors validation MSE with patience
100 and a minimum improvement of $10^{-8}$. Training terminates at epoch 368,
and the retained checkpoint is from epoch 268, where validation MSE is
0.012336. The head is trained on CPU.

\begin{table}[t]
\centering
\caption{Prediction accuracy and rank correlation of the retained
four-feature STAS head on the validation and held-out test splits.}
\label{tab:appendix_stas_validation}
\setlength{\tabcolsep}{5.5pt}
\begin{tabular}{lrrrrr}
\toprule
Split & $N$ & MAE$\downarrow$ & RMSE$\downarrow$ & PCC$\uparrow$ & SRCC$\uparrow$ \\
\midrule
Validation & 50 & 0.0930 & 0.1111 & 0.2914 & 0.3424 \\
Test & 51 & 0.0810 & 0.1003 & 0.6147 & 0.7184 \\
\bottomrule
\end{tabular}
\end{table}

\section{Evaluation Metrics}
\label{app:metrics}

\subsection{Content-Preservation Metrics}
\label{app:preservation_metrics}

All content-preservation metrics compare the source with the final
$512\times512$ RGB tile, represented as float32 values in $[0,1]$; the
$1\times3$ visualization is not evaluated. With $N=3HW$, MAE and RMSE are
\begin{align}
\mathrm{MAE}(I,\hat I)&=\frac{1}{N}\sum_p |I_p-\hat I_p|,\\
\mathrm{RMSE}(I,\hat I)&=
\sqrt{\frac{1}{N}\sum_p(I_p-\hat I_p)^2}.
\end{align}

SSIM is computed on RGB images with
\path{skimage.metrics.structural_similarity}, using
\path{channel_axis=-1} and \path{data_range=1.0}. Under scikit-image 0.25.2,
the remaining defaults correspond to a $7\times7$ uniform window,
$K_1=0.01$, $K_2=0.03$, and sample covariance; the three channel scores are
then averaged. LPIPS 0.1.4 uses the pretrained AlexNet backbone with
\path{normalize=True}, which maps $[0,1]$ tensors to the expected input range.
Full-image DISTS uses
\path{DISTS_pytorch.DISTS(load_weights=True)} version 0.1 on tensors of shape
$1\times3\times512\times512$. We compute every metric per image and report
the arithmetic mean over the benchmark.

\subsection{Horizontal Seam Metrics}
\label{app:seam_metrics}

Tiling Score follows the retained Tiled Diffusion implementation. For an
output tile $X\in[0,1]^{H\times W\times3}$ and offset $w=15$, define
\begin{align}
d_0 &= \operatorname{mean}|X[:,W-1,:]-X[:,0,:]|,\\
d_L &= \operatorname{mean}|X[:,W-w-1,:]-X[:,W-w,:]|,\\
d_R &= \operatorname{mean}|X[:,w+1,:]-X[:,w,:]|,\\
\mathrm{TS}(X)&=\max(d_0,d_L,d_R).
\label{eq:appendix_ts}
\end{align}
TS compares the periodic boundary with two nearby one-pixel RGB differences
and returns the largest of the three probes; lower values indicate a smoother
connection.

For Gradient Seam Error, the output is converted to luma using the
coefficients in Section \ref{app:stas_features}. We compute the $3\times3$
Sobel magnitude $G=\mathcal{G}(Y)$ without further normalization. For seam
band width $s=16$,
\begin{equation}
\mathrm{GSE}(X)=\frac{1}{Hs}
\left\|G[:,0{:}s]-G[:,W-s{:}W]\right\|_1.
\label{eq:appendix_gse}
\end{equation}
The two bands are compared at corresponding rows and within-band offsets.
Both TS and GSE evaluate only the output's periodic left--right connection and
do not depend on the source image.

\section{Additional Experiments}
\label{app:additional_experiments}

\subsection{Complete Depth-Fusion-Strength Sweep}
\label{app:alpha_sweep}

Table \ref{tab:appendix_alpha_sweep} aggregates 7,035 successful per-image
records, covering 21 values of $\alpha_D$ at increments of 0.05 over the same
335 inputs. All other settings remain fixed. Relative to $\alpha_D=0$, full
guidance reduces MAE by 12.79\%, LPIPS by 26.72\%, TS by 8.47\%, and GSE by
11.24\%, while increasing SSIM by 5.35\%. The setting $\alpha_D=1$ is best
for all reported metrics except GSE, whose minimum occurs at
$\alpha_D=0.95$ (0.30680 versus 0.30743). TS varies non-monotonically at
intermediate strengths, indicating that the preservation--seam trade-off is
not governed by a monotonic objective.

\begin{table*}[t]
\centering
\scriptsize
\setlength{\tabcolsep}{3.1pt}
\caption{Complete depth-fusion-strength sweep on the 335-image benchmark. Values are mean $\pm$ population standard deviation over the same 335 inputs.}
\label{tab:appendix_alpha_sweep}
\begin{tabular}{rccccccc}
\toprule
$\alpha_D$ & MAE$\downarrow$ & RMSE$\downarrow$ & SSIM$\uparrow$ & LPIPS$\downarrow$ & DISTS$\downarrow$ & TS$\downarrow$ & GSE$\downarrow$ \\
\midrule
0.00 & 0.05079 $\pm$ 0.01787 & 0.08557 $\pm$ 0.02579 & 0.66219 $\pm$ 0.11733 & 0.11338 $\pm$ 0.03934 & 0.08043 $\pm$ 0.02095 & 0.06705 $\pm$ 0.03817 & 0.34638 $\pm$ 0.14384 \\
0.05 & 0.05046 $\pm$ 0.01785 & 0.08489 $\pm$ 0.02578 & 0.66349 $\pm$ 0.11741 & 0.11143 $\pm$ 0.03935 & 0.07904 $\pm$ 0.02013 & 0.06542 $\pm$ 0.03686 & 0.34048 $\pm$ 0.14327 \\
0.10 & 0.05017 $\pm$ 0.01780 & 0.08427 $\pm$ 0.02570 & 0.66469 $\pm$ 0.11742 & 0.11001 $\pm$ 0.03917 & 0.07846 $\pm$ 0.02020 & 0.06497 $\pm$ 0.03694 & 0.33459 $\pm$ 0.14191 \\
0.15 & 0.04996 $\pm$ 0.01775 & 0.08384 $\pm$ 0.02563 & 0.66582 $\pm$ 0.11748 & 0.10871 $\pm$ 0.03912 & 0.07807 $\pm$ 0.02001 & 0.06464 $\pm$ 0.03690 & 0.33145 $\pm$ 0.14110 \\
0.20 & 0.04975 $\pm$ 0.01771 & 0.08337 $\pm$ 0.02557 & 0.66680 $\pm$ 0.11744 & 0.10779 $\pm$ 0.03898 & 0.07748 $\pm$ 0.01973 & 0.06448 $\pm$ 0.03711 & 0.32671 $\pm$ 0.14150 \\
0.25 & 0.04955 $\pm$ 0.01767 & 0.08294 $\pm$ 0.02553 & 0.66793 $\pm$ 0.11735 & 0.10664 $\pm$ 0.03894 & 0.07723 $\pm$ 0.01965 & 0.06519 $\pm$ 0.03881 & 0.32369 $\pm$ 0.14135 \\
0.30 & 0.04927 $\pm$ 0.01758 & 0.08238 $\pm$ 0.02542 & 0.66952 $\pm$ 0.11697 & 0.10505 $\pm$ 0.03845 & 0.07654 $\pm$ 0.01950 & 0.06512 $\pm$ 0.03977 & 0.32101 $\pm$ 0.14076 \\
0.35 & 0.04901 $\pm$ 0.01750 & 0.08185 $\pm$ 0.02532 & 0.67095 $\pm$ 0.11703 & 0.10404 $\pm$ 0.03833 & 0.07585 $\pm$ 0.01926 & 0.06381 $\pm$ 0.03924 & 0.31451 $\pm$ 0.13778 \\
0.40 & 0.04878 $\pm$ 0.01745 & 0.08142 $\pm$ 0.02526 & 0.67213 $\pm$ 0.11698 & 0.10298 $\pm$ 0.03812 & 0.07545 $\pm$ 0.01922 & 0.06394 $\pm$ 0.03873 & 0.31097 $\pm$ 0.13664 \\
0.45 & 0.04850 $\pm$ 0.01735 & 0.08090 $\pm$ 0.02512 & 0.67381 $\pm$ 0.11677 & 0.10163 $\pm$ 0.03775 & 0.07484 $\pm$ 0.01915 & 0.06623 $\pm$ 0.04222 & 0.30864 $\pm$ 0.13572 \\
0.50 & 0.04817 $\pm$ 0.01724 & 0.08030 $\pm$ 0.02499 & 0.67561 $\pm$ 0.11634 & 0.09998 $\pm$ 0.03702 & 0.07380 $\pm$ 0.01882 & 0.06717 $\pm$ 0.04324 & 0.30871 $\pm$ 0.13599 \\
0.55 & 0.04782 $\pm$ 0.01714 & 0.07970 $\pm$ 0.02485 & 0.67751 $\pm$ 0.11608 & 0.09832 $\pm$ 0.03646 & 0.07303 $\pm$ 0.01861 & 0.06849 $\pm$ 0.04366 & 0.30857 $\pm$ 0.13553 \\
0.60 & 0.04752 $\pm$ 0.01710 & 0.07917 $\pm$ 0.02480 & 0.67896 $\pm$ 0.11598 & 0.09692 $\pm$ 0.03607 & 0.07246 $\pm$ 0.01837 & 0.06928 $\pm$ 0.04344 & 0.30836 $\pm$ 0.13633 \\
0.65 & 0.04721 $\pm$ 0.01703 & 0.07863 $\pm$ 0.02470 & 0.68053 $\pm$ 0.11568 & 0.09559 $\pm$ 0.03556 & 0.07184 $\pm$ 0.01824 & 0.06886 $\pm$ 0.04249 & 0.30823 $\pm$ 0.13631 \\
0.70 & 0.04687 $\pm$ 0.01695 & 0.07807 $\pm$ 0.02459 & 0.68245 $\pm$ 0.11539 & 0.09414 $\pm$ 0.03520 & 0.07108 $\pm$ 0.01814 & 0.06866 $\pm$ 0.04131 & 0.30856 $\pm$ 0.13650 \\
0.75 & 0.04651 $\pm$ 0.01686 & 0.07746 $\pm$ 0.02445 & 0.68436 $\pm$ 0.11510 & 0.09249 $\pm$ 0.03464 & 0.07021 $\pm$ 0.01785 & 0.06800 $\pm$ 0.04031 & 0.30815 $\pm$ 0.13602 \\
0.80 & 0.04610 $\pm$ 0.01676 & 0.07679 $\pm$ 0.02431 & 0.68662 $\pm$ 0.11468 & 0.09069 $\pm$ 0.03405 & 0.06929 $\pm$ 0.01771 & 0.06808 $\pm$ 0.04015 & 0.30793 $\pm$ 0.13586 \\
0.85 & 0.04570 $\pm$ 0.01666 & 0.07616 $\pm$ 0.02418 & 0.68891 $\pm$ 0.11420 & 0.08901 $\pm$ 0.03356 & 0.06828 $\pm$ 0.01741 & 0.06673 $\pm$ 0.03809 & 0.30779 $\pm$ 0.13548 \\
0.90 & 0.04524 $\pm$ 0.01655 & 0.07543 $\pm$ 0.02403 & 0.69170 $\pm$ 0.11356 & 0.08696 $\pm$ 0.03283 & 0.06711 $\pm$ 0.01716 & 0.06551 $\pm$ 0.03675 & 0.30705 $\pm$ 0.13553 \\
0.95 & 0.04471 $\pm$ 0.01640 & 0.07462 $\pm$ 0.02384 & 0.69505 $\pm$ 0.11277 & 0.08476 $\pm$ 0.03206 & 0.06561 $\pm$ 0.01673 & 0.06342 $\pm$ 0.03480 & 0.30680 $\pm$ 0.13523 \\
1.00 & 0.04429 $\pm$ 0.01632 & 0.07398 $\pm$ 0.02373 & 0.69759 $\pm$ 0.11232 & 0.08308 $\pm$ 0.03150 & 0.06481 $\pm$ 0.01655 & 0.06137 $\pm$ 0.03280 & 0.30743 $\pm$ 0.13508 \\
\bottomrule
\end{tabular}
\end{table*}

\subsection{Depth-Shaping Exponent}
\label{app:gamma_sensitivity}

Table \ref{tab:appendix_gamma_sensitivity} reports the retained gamma runs
for $\alpha_D\in\{0.2,1.0\}$. These experiments use Depth Anything V2
Small rather than the ZoeDepth cache used in the main depth-strength sweep;
the $\alpha_D=1,\gamma=0.5$ result is therefore not expected to match the
main ZoeDepth result exactly. For $\alpha_D=0.2$, $\gamma=0.35$ achieves the
lowest MAE (0.05001), whereas $\gamma=0.50$ achieves the lowest TS (0.06489).
For full guidance, $\gamma=0.35$ gives the lowest MAE (0.04604), while
$\gamma=1.00$ gives the lowest TS (0.06597). These differences are descriptive
because no significance test was retained for this sweep.

\begin{table*}[t]
\centering
\scriptsize
\setlength{\tabcolsep}{3.1pt}
\caption{Sensitivity to the depth-shaping exponent $\gamma$. Values are mean $\pm$ population standard deviation over 335 images.}
\label{tab:appendix_gamma_sensitivity}
\begin{tabular}{rrccccccc}
\toprule
$\alpha_D$ & $\gamma$ & MAE$\downarrow$ & RMSE$\downarrow$ & SSIM$\uparrow$ & LPIPS$\downarrow$ & DISTS$\downarrow$ & TS$\downarrow$ & GSE$\downarrow$ \\
\midrule
0.20 & 0.35 & 0.05001 $\pm$ 0.01776 & 0.08385 $\pm$ 0.02564 & 0.66563 $\pm$ 0.11751 & 0.10921 $\pm$ 0.03912 & 0.07829 $\pm$ 0.02029 & 0.06498 $\pm$ 0.03767 & 0.33047 $\pm$ 0.14049 \\
0.20 & 0.50 & 0.05011 $\pm$ 0.01777 & 0.08407 $\pm$ 0.02566 & 0.66518 $\pm$ 0.11746 & 0.10964 $\pm$ 0.03893 & 0.07834 $\pm$ 0.02023 & 0.06489 $\pm$ 0.03744 & 0.33226 $\pm$ 0.14079 \\
0.20 & 0.75 & 0.05022 $\pm$ 0.01778 & 0.08432 $\pm$ 0.02567 & 0.66473 $\pm$ 0.11744 & 0.11032 $\pm$ 0.03886 & 0.07850 $\pm$ 0.02018 & 0.06526 $\pm$ 0.03721 & 0.33480 $\pm$ 0.14119 \\
0.20 & 1.00 & 0.05031 $\pm$ 0.01781 & 0.08451 $\pm$ 0.02572 & 0.66431 $\pm$ 0.11745 & 0.11081 $\pm$ 0.03891 & 0.07885 $\pm$ 0.02017 & 0.06576 $\pm$ 0.03750 & 0.33675 $\pm$ 0.14164 \\
1.00 & 0.35 & 0.04604 $\pm$ 0.01681 & 0.07670 $\pm$ 0.02452 & 0.68766 $\pm$ 0.11516 & 0.09120 $\pm$ 0.03462 & 0.06838 $\pm$ 0.01800 & 0.06636 $\pm$ 0.03872 & 0.30763 $\pm$ 0.13399 \\
1.00 & 0.50 & 0.04676 $\pm$ 0.01699 & 0.07794 $\pm$ 0.02475 & 0.68355 $\pm$ 0.11585 & 0.09428 $\pm$ 0.03560 & 0.07018 $\pm$ 0.01850 & 0.06703 $\pm$ 0.03989 & 0.30910 $\pm$ 0.13476 \\
1.00 & 0.75 & 0.04757 $\pm$ 0.01722 & 0.07938 $\pm$ 0.02507 & 0.67912 $\pm$ 0.11653 & 0.09807 $\pm$ 0.03670 & 0.07227 $\pm$ 0.01925 & 0.06641 $\pm$ 0.03931 & 0.31198 $\pm$ 0.13521 \\
1.00 & 1.00 & 0.04812 $\pm$ 0.01737 & 0.08040 $\pm$ 0.02528 & 0.67619 $\pm$ 0.11687 & 0.10079 $\pm$ 0.03740 & 0.07344 $\pm$ 0.01906 & 0.06597 $\pm$ 0.03961 & 0.31542 $\pm$ 0.13606 \\
\bottomrule
\end{tabular}
\end{table*}

\subsection{Semantic Category Analysis}
\label{app:category_analysis}

For metric $m$, Tables \ref{tab:appendix_category_preservation} and
\ref{tab:appendix_category_seam} report
\[
\Delta m=\overline{m}_{\alpha_D=1}-\overline{m}_{\alpha_D=0},
\qquad
\Delta m_{\mathrm{rel}}=
100\Delta m/\overline{m}_{\alpha_D=0}.
\]
Negative changes are favorable for lower-is-better metrics, whereas positive
changes are favorable for SSIM. DepthTile improves all five preservation
metrics and GSE in every semantic category, and improves TS in 11 of 12
categories. Animals is the only exception: TS increases from 0.04513 to
0.04619 (+2.35\%), although all preservation metrics and GSE improve. The
largest relative TS reductions occur for Desert (-24.46\%), Object Foreground
(-17.79\%), and Beach (-17.60\%); Forest changes only marginally (-0.20\%).

\begin{table*}[t]
\centering
\scriptsize
\setlength{\tabcolsep}{4.0pt}
\caption{Category-wise change from Tiled Diffusion ($\alpha_D=0$) to DepthTile ($\alpha_D=1$) for content-preservation metrics. Each cell gives the absolute change followed by the relative change in parentheses.}
\label{tab:appendix_category_preservation}
\begin{tabular}{lrrrrrr}
\toprule
Category & $N$ & $\Delta$MAE & $\Delta$RMSE & $\Delta$SSIM & $\Delta$LPIPS & $\Delta$DISTS \\
\midrule
animals & 32 & -0.00614 (-14.0\%) & -0.01140 (-14.9\%) & +0.03213 (+4.4\%) & -0.02731 (-29.1\%) & -0.01134 (-20.6\%) \\
architecture & 41 & -0.00664 (-13.0\%) & -0.01186 (-13.6\%) & +0.03598 (+5.2\%) & -0.02775 (-27.9\%) & -0.01541 (-19.4\%) \\
beach & 26 & -0.00573 (-13.1\%) & -0.01080 (-14.2\%) & +0.03325 (+4.9\%) & -0.02827 (-27.5\%) & -0.01467 (-17.9\%) \\
desert & 17 & -0.00536 (-12.3\%) & -0.00980 (-13.5\%) & +0.03276 (+4.8\%) & -0.02823 (-27.0\%) & -0.02293 (-23.5\%) \\
forest & 24 & -0.00824 (-11.4\%) & -0.01351 (-12.0\%) & +0.04606 (+8.3\%) & -0.03794 (-24.0\%) & -0.01738 (-18.1\%) \\
garden & 30 & -0.00749 (-11.0\%) & -0.01273 (-11.6\%) & +0.03754 (+6.8\%) & -0.03496 (-24.6\%) & -0.01500 (-17.1\%) \\
indoor & 31 & -0.00610 (-17.7\%) & -0.01169 (-18.5\%) & +0.03082 (+3.9\%) & -0.02291 (-29.5\%) & -0.01209 (-19.2\%) \\
landscape & 33 & -0.00686 (-12.3\%) & -0.01224 (-13.2\%) & +0.03788 (+6.3\%) & -0.03264 (-24.8\%) & -0.01811 (-19.7\%) \\
mountain & 29 & -0.00685 (-11.7\%) & -0.01200 (-12.4\%) & +0.03541 (+5.8\%) & -0.03283 (-23.6\%) & -0.01595 (-16.6\%) \\
object\_foreground & 35 & -0.00395 (-10.8\%) & -0.00740 (-11.2\%) & +0.02251 (+3.0\%) & -0.02084 (-26.5\%) & -0.00976 (-15.5\%) \\
road & 17 & -0.00705 (-13.6\%) & -0.01257 (-14.5\%) & +0.04066 (+6.7\%) & -0.04022 (-29.4\%) & -0.02537 (-26.5\%) \\
street & 20 & -0.00856 (-15.7\%) & -0.01447 (-15.9\%) & +0.04947 (+7.7\%) & -0.04050 (-31.8\%) & -0.01973 (-24.8\%) \\
\bottomrule
\end{tabular}
\end{table*}

\begin{table}[t]
\centering
\scriptsize
\setlength{\tabcolsep}{4.0pt}
\caption{Category-wise change from Tiled Diffusion ($\alpha_D=0$) to DepthTile ($\alpha_D=1$) for seam-quality metrics. Each cell gives the absolute change followed by the relative change in parentheses.}
\label{tab:appendix_category_seam}
\begin{tabular}{lrrr}
\toprule
Category & $N$ & $\Delta$TS & $\Delta$GSE \\
\midrule
animals & 32 & +0.00106 (+2.3\%) & -0.02988 (-10.5\%) \\
architecture & 41 & -0.00350 (-5.2\%) & -0.04242 (-11.4\%) \\
beach & 26 & -0.01107 (-17.6\%) & -0.03044 (-10.5\%) \\
desert & 17 & -0.01736 (-24.5\%) & -0.03967 (-13.5\%) \\
forest & 24 & -0.00017 (-0.2\%) & -0.04301 (-9.5\%) \\
garden & 30 & -0.00097 (-1.2\%) & -0.04881 (-11.2\%) \\
indoor & 31 & -0.00196 (-3.1\%) & -0.04346 (-12.4\%) \\
landscape & 33 & -0.00707 (-10.5\%) & -0.05211 (-14.2\%) \\
mountain & 29 & -0.00848 (-11.0\%) & -0.02084 (-5.9\%) \\
object\_foreground & 35 & -0.01017 (-17.8\%) & -0.02550 (-11.2\%) \\
road & 17 & -0.00409 (-7.2\%) & -0.04288 (-12.8\%) \\
street & 20 & -0.01057 (-13.2\%) & -0.05479 (-13.3\%) \\
\bottomrule
\end{tabular}
\end{table}

\subsection{Computational Cost}
\label{app:efficiency}

We measure computational cost at batch size one and $512\times512$
resolution on an NVIDIA A100-SXM4-40GB. Synchronized means over 335 images
exclude one-time model loading, image serialization, and metric evaluation.
ZoeDepth-N requires $0.2477\pm0.0435$ s/image, SDXL generation requires
$3.4500\pm0.3169$ s/image, and the complete pipeline requires
$3.6977\pm0.3185$ s/image. The conservative idle-subtracted peak memory is
15.23 GiB.

The STAS MLP contains 2,433 parameters and 2,336 linear MACs, corresponding to
4,672 FLOPs when multiplication and addition are counted separately; bias and
activation operations are excluded. After 5,000 warm-up passes, seven trials
of 50,000 batch-one FP32 CUDA forward passes yield a median head-only latency
of 0.10894 ms/image.

\section{Construction and Adaptation of Comparison Methods}
\label{app:baseline_adaptation}

We adapt four diffusion-based methods to a common horizontal self-tiling
protocol: given a $512\times512$ image $I$, each method produces one
$512\times512$ tile $\widetilde I$ whose left and right boundaries should
repeat seamlessly. The $1\times3$ result is formed only after inference as
$[\widetilde I\mid\widetilde I\mid\widetilde I]$ and is not used by the
preservation metrics. Across the 335 inputs, the external baselines process
images in a deterministic order, with seed $i-1$ assigned to input $i$. No
baseline is trained or fine-tuned on the benchmark.

\subsection{Seamless Tile Inpainting}
\label{app:baseline_sti}

Standard inpainting cannot jointly process opposite image boundaries because
they are spatially separated. We therefore adapt Seamless Tile Inpainting
(STI) through a shift--inpaint--unshift construction. Let
$\mathcal R_{\delta}$ denote a circular horizontal shift by $\delta$ pixels,
and let $C_{96}$ denote a 96-pixel-wide vertical mask centered at $x=W/2$.
The adapted result is
\[
\widetilde I =
\mathcal R_{-W/2}\!\left(
  \operatorname{Inpaint}\!\left(\mathcal R_{W/2}(I),C_{96}\right)
\right).
\]
The half-width shift moves the original left--right junction to the image
center, allowing a standard inpainting model to process both sides within one
local region. The inverse shift restores the original coordinates. Compared
with masking two disconnected outer borders, this adaptation confines editing
to the junction neighborhood.

We use the Stable Diffusion 1.5 inpainting checkpoint
\path{runwayml/}\allowbreak\path{stable-diffusion-inpainting}, 30 inference steps, inpainting
strength 1.0, and classifier-free guidance scale 7.5. The positive prompt is ``preserve the original image, seamless texture,
coherent details,'' and the negative prompt is ``visible seam, border, frame,
blurry, distorted, text, watermark.''

\subsection{Asymmetric Tiling}
\label{app:baseline_at}

Asymmetric Tiling (AT) modifies U-Net convolutional padding so that opposite
boundaries exchange context during denoising. For horizontal self-tiling, we
enable circular padding only along the horizontal axis and retain the default
vertical padding, avoiding an unnecessary top--bottom periodic constraint.
Unlike STI, AT has no spatial edit mask, so img2img denoising may update the
entire $512\times512$ input.

We use \path{runwayml/stable-diffusion-v1-5}, img2img strength 0.65, 40
inference steps, an empty prompt, and guidance scale 1.0. The empty prompt and
unit guidance minimize semantic steering, isolating the effect of horizontal
circular U-Net padding.

\subsection{MultiDiffusion}
\label{app:baseline_multidiffusion}

MultiDiffusion coordinates overlapping denoising views over a large canvas.
Our adaptation encodes $I$ into latent $z_I$ and constructs a three-period
horizontal canvas by repeating both the source latent and its initial noise:
\[
z_I^{(3)}=[z_I\mid z_I\mid z_I],
\qquad
\epsilon^{(3)}=[\epsilon\mid\epsilon\mid\epsilon].
\]
Overlapping views are denoised and fused on the $1536\times512$ image-space
canvas, with circular handling along the horizontal axis. After denoising, we
average corresponding latent coordinates from the three periods. Let $W_z$
denote the width of one latent period:
\[
\bar z(x,y)=\frac{1}{3}\sum_{j=0}^{2}z^{(3)}(x+jW_z,y),
\]
The averaged latent $\bar z$ is decoded into one output tile. Repeating the
same source and noise removes period-dependent initialization differences,
periodic view handling supplies context across the outer boundary, and
averaging returns a standard-size tile rather than a panorama.

We use \path{runwayml/stable-diffusion-v1-5} with DDIM sampling. The base
schedule contains 30 steps and img2img strength 0.60 selects its final 18
denoising timesteps. Guidance scale is 5.0 and view batch size is one. The positive prompt is ``preserve the original scene, seamless periodic
image, realistic photo, consistent lighting,'' and the negative prompt is
``visible seam, blurry, distorted, duplicate objects, text, watermark.''

\subsection{SpotDiffusion}
\label{app:baseline_spotdiffusion}

SpotDiffusion uses the same three-period source-latent and noise construction
but replaces overlapping-view fusion with temporally moving, non-overlapping
windows. At each timestep, the U-Net processes a partition of latent windows,
after which the horizontal origin changes so that locations on a window
boundary at one step move into a window interior at later steps. Following the
final update, we undo the accumulated phase offset and average corresponding
latent periods to obtain one tile. This moving partition distributes
window-boundary artifacts over time while preserving three-period horizontal
context.

We use \path{runwayml/stable-diffusion-v1-5} and DDIM. Each latent window is
$64\times64$, corresponding to a $512\times512$ image-space view on the
$1536\times512$ canvas. The base schedule contains 30 steps; img2img strength
0.35 selects its final 10 denoising timesteps. Guidance scale is 5.0, and the
positive and negative prompts are the same as for MultiDiffusion. The
per-timestep horizontal offsets are generated deterministically from the
image-specific seed.

\subsection{Tiled Diffusion and Evaluation Controls}
\label{app:baseline_controls}

The Tiled Diffusion baseline is obtained by setting $\alpha_D=0$ in the same
SDXL runner used by DepthTile. This exactly recovers the fixed base mask while
holding the prompt, scheduler, denoising steps, context width, and latent
wrapping constant. It is therefore the controlled same-backbone comparison.
By contrast, STI, AT, MultiDiffusion, and SpotDiffusion are task-adapted
external baselines built on Stable Diffusion 1.5, and this backbone difference
should be considered when interpreting their results.

The retained experiment record specifies the constructions and numerical
settings above, and the archive contains the corresponding per-image metrics.
The current release does not yet include executable adapters, exact upstream
revisions, or licenses for the four external baselines. It also lacks
MultiDiffusion's view size and stride and SpotDiffusion's exact offset-sampling
rule. Consequently, the descriptions above document the settings associated
with the frozen results, but do not support one-command regeneration of the
external-baseline rows. These implementation and licensing details should be
restored before public release.

\section{Vertical Tiling Status}
\label{app:vertical_status}

Our implementation and evaluation are restricted to horizontal self-tiling,
and no controlled 335-image vertical comparison is available. Although
\path{tiled-diffusion/}\allowbreak\path{utils.py} provides \path{pad_tensor_y} and
\path{wrap_edges_y}, the active Differential Diffusion pipeline applies
padding, wrapping, cropping, and evaluation only along the horizontal axis;
the corresponding vertical calls are disabled. The retained depth prior and
base mask are also horizontal. Rotating the inputs or reusing horizontal
outputs would not constitute a controlled vertical evaluation. A valid
extension requires a vertical protection mask, top--bottom depth padding,
active y-axis latent wrapping, a verified vertical seam metric, and a new
paired $\alpha_D=0$ versus $\alpha_D=1$ evaluation over all 335 inputs.
Accordingly, we do not report vertical results.

\section{Additional Qualitative Comparisons}
\enlargethispage{2\baselineskip}
\label{app:additional_qualitative}

\begin{figure*}[p]
\centering
\includegraphics[width=0.94\textwidth,height=0.85\textheight,keepaspectratio]{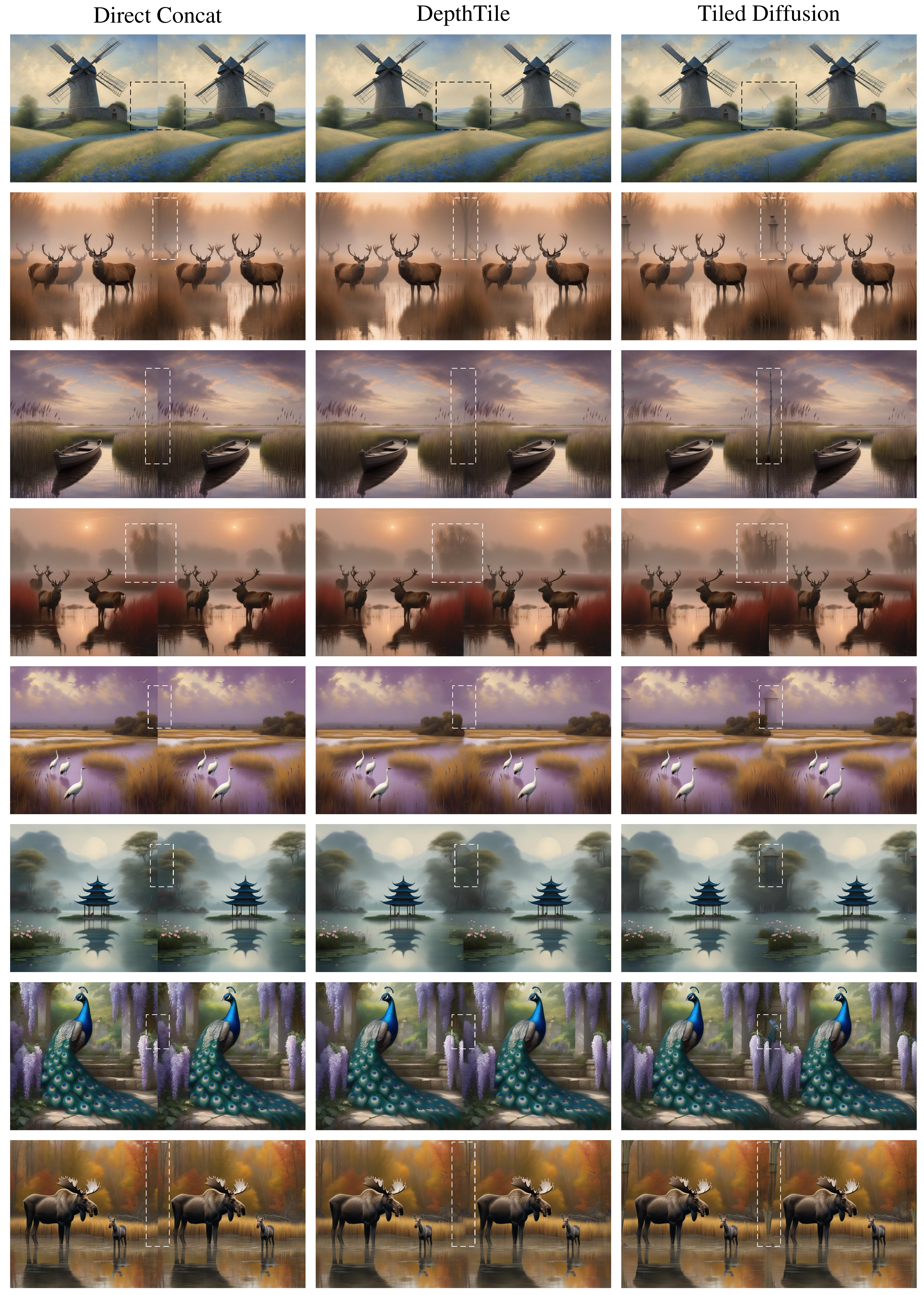}
\caption{Artifact suppression in existing-image self-tiling. Each row compares
$1\times2$ repetitions from direct concatenation, DepthTile, and Tiled
Diffusion. Black or white dashed boxes mark artifacts introduced by Tiled
Diffusion but avoided by DepthTile.}
\label{fig:app_artifacts}
\end{figure*}

\begin{figure*}[p]
\centering
\includegraphics[width=\textwidth,height=0.85\textheight,keepaspectratio]{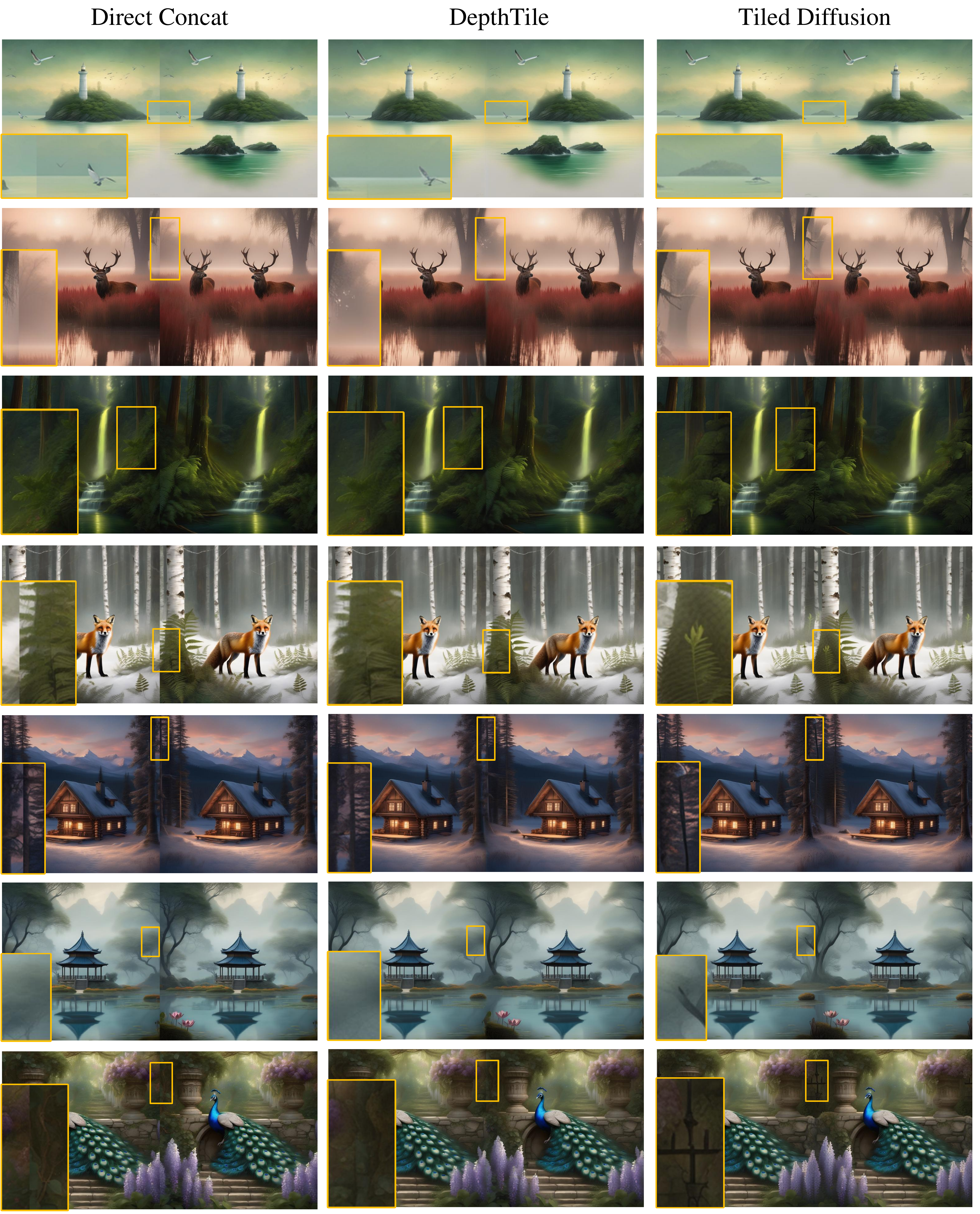}
\caption{Fine-detail preservation. Yellow boxes identify the compared regions,
and the lower-left insets enlarge the same crop for all methods. DepthTile
retains local structures closer to the direct-concatenation reference than
Tiled Diffusion.}
\label{fig:app_details}
\end{figure*}

\begin{figure*}[p]
\centering
\includegraphics[width=\textwidth,height=0.85\textheight,keepaspectratio]{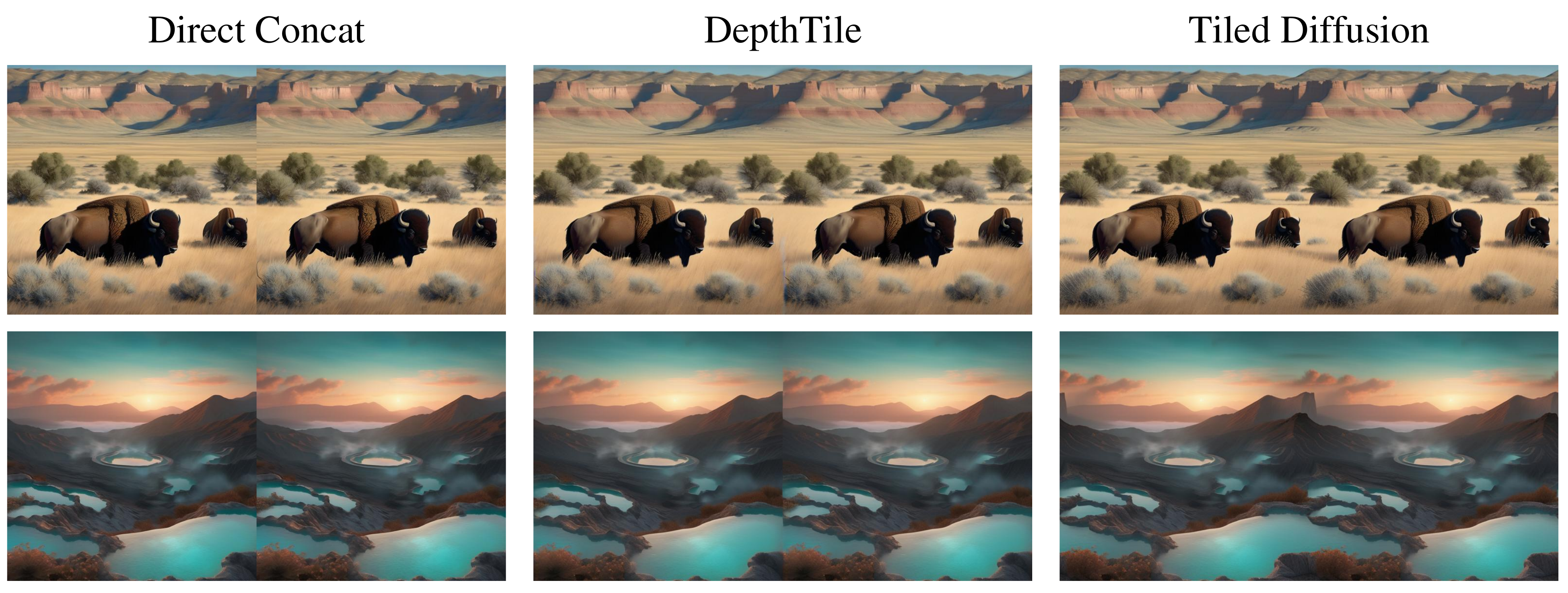}
\caption{Examples in which stronger content preservation leaves a more visible
center seam. DepthTile restricts edits to foreground structures and may
therefore retain a slightly stronger discontinuity than Tiled Diffusion.}
\label{fig:app_depthtile_seams}
\end{figure*}

\begin{figure*}[p]
\centering
\includegraphics[width=\textwidth,height=0.85\textheight,keepaspectratio]{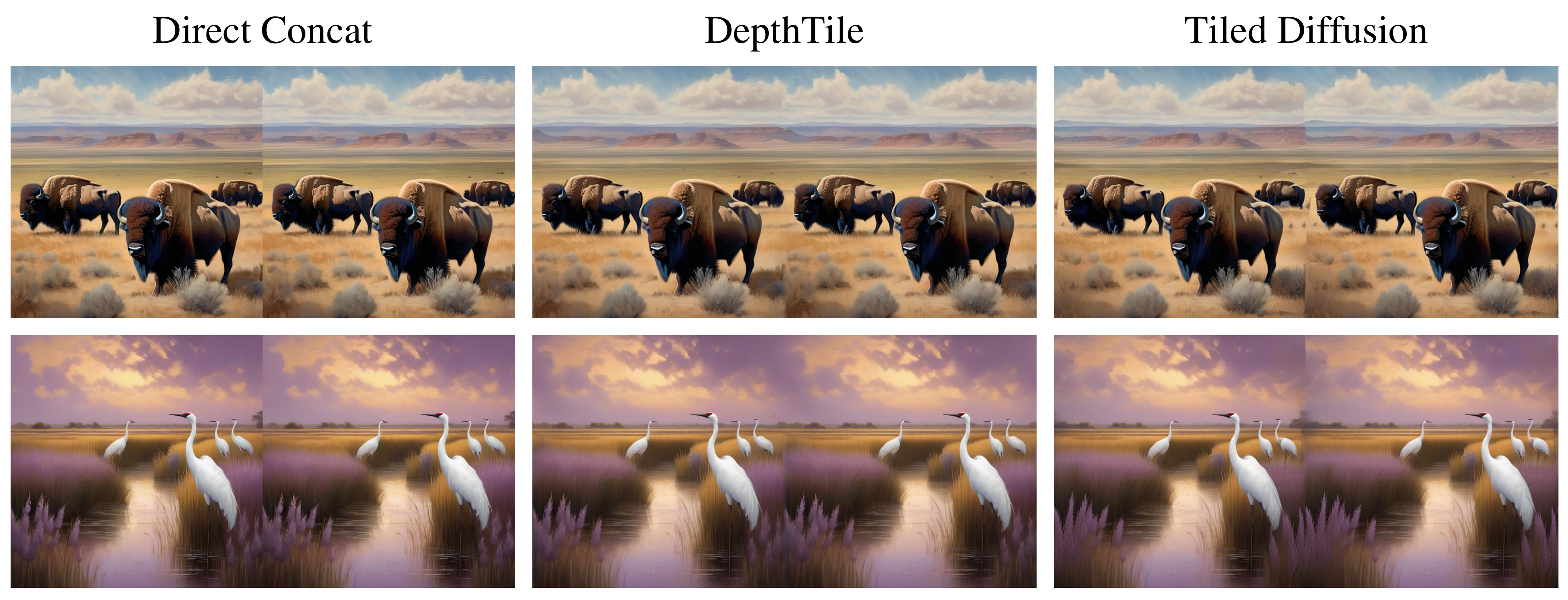}
\caption{Examples in which Tiled Diffusion produces the more visible center
seam. Aggressive boundary regeneration can increase cross-seam structural or
appearance inconsistency, whereas DepthTile preserves a more coherent
transition.}
\label{fig:app_td_seams}
\end{figure*}

Each row in Figs.~\ref{fig:app_artifacts}--\ref{fig:app_td_seams}
corresponds to the same source image and compares $1\times2$ horizontal
repetitions produced by direct concatenation, DepthTile, and Tiled Diffusion.
Direct concatenation serves as the unchanged-content reference.

\paragraph{Artifact suppression.}
Figure~\ref{fig:app_artifacts} highlights artifacts generated by Tiled
Diffusion but absent from both the reference and DepthTile. The dashed boxes
show that unrestricted boundary regeneration can introduce structures not
supported by the input, whereas depth-guided protection suppresses these
changes.

\paragraph{Fine-detail preservation.}
Figure~\ref{fig:app_details} compares local structures marked by yellow boxes.
The lower-left inset in each result enlarges the same region. Tiled Diffusion
often modifies or removes thin objects and fine textures, while DepthTile
retains details closer to the direct-concatenation reference.

\paragraph{Seam behavior.}
The qualitative results also illustrate the preservation--tileability
trade-off. In some cases, DepthTile leaves a more visible center seam than
Tiled Diffusion because stronger foreground protection limits the available
boundary edits (Fig.~\ref{fig:app_depthtile_seams}). In other cases, however,
the stronger regeneration of Tiled Diffusion introduces a larger structural
or appearance mismatch across the junction, and DepthTile produces the
smoother transition (Fig.~\ref{fig:app_td_seams}).

\clearpage

\end{document}